\documentclass{article}
\usepackage{iclr2027_conference,times}
\iclrfinalcopy

\usepackage{amsmath,amssymb,amsfonts}
\usepackage{booktabs}
\usepackage{longtable}
\usepackage{graphicx}
\usepackage{microtype}
\usepackage{url}
\usepackage[table]{xcolor}
\usepackage{siunitx}
\usepackage[hidelinks]{hyperref}
\usepackage{caption}
\usepackage{fvextra}
\newcommand{\optset}{\mathcal{O}}
\newcommand{\ctxmass}{\mu}
\newcommand{\floorval}{0.5}
\newcommand{\code}[1]{\texttt{\small\hyphenchar\font=`\-\relax #1}}
\newcommand{\bpechar}[1]{{\fontencoding{T1}\selectfont #1}}
\newcommand{\sptok}{\texttt{\small\char`\_}}
\newcommand{\sq}[1]{\texttt{\small '\char`\_#1'}}   %
\newcommand{\gq}[1]{\texttt{\small '\bpechar{\.{G}}#1'}}
\newcommand{\cq}[1]{\texttt{\small '\bpechar{\.{C}}#1'}}

\title{Scoring the Wrong Question:\\Readout Failures in\\Constrained-Option Evaluation}

\author{Jiaxuan Guo$^{1}$\thanks{Corresponding author.}\hphantom{$^{*}$} \enspace Kejia Zhang$^{2}$ \enspace Shuo Xin$^{1}$ \enspace
Jingxin Yang$^{1}$ \enspace Youran Sun$^{2}$ \enspace Haizhao Yang$^{2}$ \\
$^{1}$Stanford University \quad $^{2}$University of Maryland, College Park \\
\texttt{guojx@stanford.edu}}

\begin{document}

\maketitle
\lhead{}\renewcommand{\headrulewidth}{0pt}

\begin{abstract}
Constrained-option scoring reads a model's probabilities for a fixed set of
permitted answers, so it returns a score even when the model is about to write
something else. We study a prompt that quotes a multiple-choice item, ending in
the item's own answer instruction, and asks for a forecast of whether a reader
model given only a short note will answer it correctly. The model can begin
answering the quoted item: on three Qwen3 checkpoints an answer letter is the
most probable token on nearly every passage, and the forecast ranks the reader's
correctness indistinguishably from chance. Beyond prior work's argmax check and
prefill repair, we contribute a label-free diagnostic of the declared options' support, their first-token mass in context; a deletion test
tracing the Qwen3 answer-letter start to the quoted instruction; and rendering
and tokenisation checks for dropped separators and coinciding first tokens.
Prefilling an answer stem lifts the median option mass from below to above one
half on all nine checkpoints scored with this prompt and, with options and
renormalisation unchanged, raises the Qwen3-averaged AUC from 0.4929 to 0.6152.
lm-polygraph's default P(True) estimator shows a related failure: it reads True
where Qwen3 without thinking favours an answer letter (MMLU) or the prompt's
(A)/(B) label (TriviaQA), and its Qwen3-averaged AUROC is indistinguishable from
chance on MMLU and below chance on TriviaQA. Scoring and renormalising the
(A)/(B) labels after an answer stem raises it to 0.632 and 0.868, renormalising
True against False in place reaches 0.660 and 0.860, and on TriviaQA both exceed
all fourteen default single-answer library estimators. The original forecast is
renormalised too, yet stays indistinguishable from chance: option mass flags low
support in both settings without labels, and only checking the intended target
shows which score still ranks.
\end{abstract}

\section{Introduction}
\label{sec:intro}

A model can answer a quoted question while an evaluation reports its
forecast of someone else's answer. In Figure~\ref{fig:failure-example}, the
prompt quotes a reading-comprehension item with a letter-answer instruction,
then asks how many of a hundred readers would answer correctly. The model
answers the item first; the scorer reads that position and returns a
distribution over its permitted numbers.

At the original position, the first token of the reply, the forecast ranks
reader correctness indistinguishably from chance: its AUC averaged over three
Qwen3 checkpoints is 0.4929 $[0.4390, 0.5458]$. On 240 shared passages, an
answer stem, text prefilled at the start of the reply, raises the
Qwen3-averaged AUC to 0.6152
and Meta-Llama-3-8B-Instruct's from 0.5231 to 0.6044.
Table~\ref{tab:consequence} gives paired intervals and Brier scores;
Section~\ref{sec:repair} compares the stemmed Brier scores with a constant
0.5 forecast.

\begin{figure}[t]
\centering
\begin{minipage}[t]{0.60\linewidth}
\small\textbf{Rendered prompt ending}\par\smallskip
\begin{Verbatim}[fontsize=\footnotesize,breaklines=true,breakanywhere=true,frame=single]
After the woman visited the United States she might feel that   _  .
A: her English was poor
B: it's easy to master English
C: it's difficult to make money
D: people there weren't very friendly
Answer with exactly one of: A, B, C, D.

Out of 100 such readers, how many answer correctly? Reply with exactly one of: 10, 30, 50, 70, 90.<|im_end|>
<|im_start|>assistant
\end{Verbatim}
\textbf{Scored here:} the next token after the final newline.
\end{minipage}\hfill
\begin{minipage}[t]{0.37\linewidth}
\small\textbf{Next-token distribution}\par\smallskip
\begin{tabular}{@{}lr@{}}
\toprule
Token & Probability \\
\midrule
\textbf{\texttt{A}} & \textbf{1.0} \\
\texttt{Answer} & $3.36\times10^{-10}$ \\
\texttt{D} & $3.28\times10^{-11}$ \\
\texttt{After} & $8.29\times10^{-12}$ \\
\texttt{5} & $5.27\times10^{-12}$ \\
\texttt{The} & $4.73\times10^{-12}$ \\
\midrule
Five-option mass & $5.36\times10^{-12}$ \\
\bottomrule
\end{tabular}
\par\medskip
\textbf{Unrestricted continuation}\par\smallskip
\fbox{\begin{minipage}{\dimexpr\linewidth-2\fboxsep-2\fboxrule\relax}
\texttt{A}\hfill answers the item\\
\texttt{50}\hfill gives the forecast
\end{minipage}}
\par\smallskip
The exact output is \texttt{A\ \ \textbackslash n50}.
\end{minipage}
\caption{The model answers the quoted item before forecasting. On this real
Qwen3-4B item, the note states the answer, A. The later forecast instruction
asks for one of five numbers, but the scored position assigns its highest
probability to A; with notes about an unrelated passage, which state no answer,
58 of 60 rows still put an answer letter first (Section~\ref{sec:failure}). The table shows the six most probable tokens and the mass
of the five options' distinct first tokens. The unrestricted continuation reaches
the forecast only after answering the item. Appendix~\ref{app:prompt} gives the complete
prompt and both scored positions; line wrapping here is for display.}
\label{fig:failure-example}
\end{figure}

\paragraph{How the scoring hides the failure.}
Constrained-option scoring compares a fixed set of legal continuations, the
\emph{declared set}, at one next-token position, the \emph{scored position}.
Renormalising over that set discards all other probability mass. Our declared
set is the forecast grid $\{10,30,50,70,90\}$, whereas the quoted item requests
A, B, C or D. Section~\ref{sec:failure} measures the answer-letter argmax
and forecast mass. The unrestricted continuation, the model's greedy output there, reveals which
instruction its first token follows.

\paragraph{What the experiments establish.}
Beyond the argmax-membership test and prefill repair of \citet{cappelletti2025prefilling}, we
add the option-mass diagnostic, localisation to the quoted instruction,
tokenisation checks and consequences in maintained evaluation software.
Deleting the quoted item's closing instruction removes the answer-letter start
on the three Qwen3 checkpoints and, at the original ending, lowers its rate to
0.016 on SmolLM2 and Mistral; Phi-3 needs the choice list removed too. Deleting only the choice list
leaves an answer-letter argmax on 67 to 89\% of items on the three Qwen3
checkpoints and SmolLM2 (Appendix~\ref{app:instruction-experiment}). Substituted instructions also
change the generated response, but the tested token-level rules do not explain how.
A judge and human labels assess instruction execution from the continuation
(Section~\ref{sec:failure}). Section~\ref{sec:families} separates answer-letter
starts from missing prompt separators across seven checkpoints in five
architecture families. lm-polygraph's
P(True) confidence ranks correctness indistinguishably from chance on MMLU \citep{hendrycks2021mmlu}
and below chance on TriviaQA \citep{joshi2017triviaqa}, averaged over three Qwen3 checkpoints without
thinking (Section~\ref{sec:failure}). The same design appears in maintained scorers:
lm-polygraph enables P(True) in 23 of its 30 configurations, AlpacaEval reads
log-probabilities over its declared set in 12 of its 44 annotator
configurations, and 27 of 131 OpenCompass LLM-judge grader configurations quote
a lettered item and read the judge's first A or B (Appendix~\ref{app:lmpolygraph}).

First-token readouts also require distinct option tokens after the actual
prompt; isolated encoding can score a shared marker. The experiment map,
Table~\ref{tab:expmap}, gives the samples and readouts;
Section~\ref{sec:repair} gives the practical checks.

\section{Related work}
\label{sec:related}

\paragraph{An option's probability and the answer it represents.}
A string can represent an answer without its probability measuring the model's
preference for that answer. \citet{holtzman-etal-2021-surface} describe how
generic strings absorb probability and leave intended answers in a poorly
calibrated tail. Their COPA-Flipped control
\citep[Section~5, Table~5]{holtzman-etal-2021-surface} removes the hypothesised
cause and the scoring-rule gap disappears; our instruction-deletion
control in Section~\ref{sec:failure} shares that design.
\citet{robinson2023leveraging} likewise distinguish sentence probability from
answer correctness. Their multiple-choice alternative requires a model to
bind a symbol to an option, and they show GPT-3 assigning its highest
probability to \code{A} under both orderings of one list. Our quoted item uses
that ability: it has its own labels and instruction, which can take over a
position scored only for forecasts. Their repair uses arbitrary symbols; our options are an ordered numeric grid
whose ordinal structure is the measurement.
\citet[Section~6]{sun2026perspectivegap} show models writing instructions for
agents in the wrong roles; here a model reads and executes an instruction
addressed to another reader.

\paragraph{Instructions inside quoted material.}
\citet[Sections~1 and~3]{greshake2023indirect} define indirect prompt injection
with an adversary placing instructions in retrieved data. The data--instruction
confusion it exploits also occurs without an attacker:
\citet[Sections~1 and~3, Definitions~3--4; Section~4]{zverev2025separate} place a probe
in the data and count its one-word answer anywhere in the output;
\citet[Sections~1 and~3.1; Table~3]{hwang2025distraction} study rewriting,
proofreading, translation and style transfer when the input itself resembles
an instruction, such as a question or a coding request.
\citet[Section~1, Figure~1; Section~3.2]{wallace2024hierarchy} assign privilege
by message type. In our scoring prompt, the quoted item's instruction and the
scorer's question share one user turn, so we infer that this hierarchy assigns
them the same privilege. These studies read generated text; we read the scorer's
side, where answering the quoted item first is enough to put its answer letter
at the scored position.

\paragraph{Prompt format and the position being scored.}
\citet{sclar2024quantifying} require instructions and options to match:
changing option characters also changes every reference to them in the
instruction. Our prompt contains two different requested option sets. Their
option-ranking metric omits probability outside the options. Both their study
and \citet{robinson2023leveraging} measure whitespace effects.
\citet{sanzguerrero2025mind} report whitespace-related accuracy changes up to
11\% and changes in model rankings. They recommend tokenising the space with
the answer letter; for numeric labels their two-element split uses the final
element, where digits remain distinct, rather than the shared first element.

\citet{wang2024mytoken} directly compare first-token scores with generated text.
Even their instruction to start with a single letter leaves misalignment.
Their subject is agreement between readouts; ours is a declared set that
excludes what sits at the scored position.
\citet{wang2024look} compare which readout survives
perturbation. An answer stem instead supplies text in the assistant response,
the distinction studied by \citet{cappelletti2025prefilling}.
\citet{karim2026attention} distinguish failure to enter answer mode from
failure to select the right candidate once in that mode, using a
log-probability margin over candidates. Our case is their gating failure:
the model's next token is not among the scored candidates. We use the existing
prefill mechanism and identify two misleading
measurements from encoding options in isolation
(Section~\ref{sec:repair}; Appendix~\ref{app:isolated-encoding}).

\paragraph{Calibration and diagnostic checks.}
Calibration and selection-bias methods operate within the scored option set
and assume that its position is correct
\citep{pmlr-v139-zhao21c,zheng2023large}; theirs is a skew within that set, ours
mass outside it. Numeric labels themselves work as a readout:
\citet{zheng2023large} report MMLU accuracy of 65.8 with numeric
labels and 67.2 with letters on \code{gpt-3.5-turbo}.
For checking metrics, \citet{shankar2024who} combine coverage and false failure
rate for boolean assertions. \citet{sai-etal-2021-perturbation} pair changes a
continuous metric must detect with invariant cases whose score must not drop.
\citet{cappelletti2025prefilling} measure the unrestricted highest-probability
token's membership in the option set across examples. Restricting the argmax
to the set would remove membership as a diagnostic while leaving option mass
measurable.

\section{Task, measurements and experiments}
\label{sec:protocol}

A writer model compresses a passage into a short note under a word budget. A reader model sees only
the note and answers the passage's multiple-choice question. The forecasting
model sees the note and the complete item and selects from
$\optset=\{\code{10},\code{30},\code{50},\code{70},\code{90}\}$:
how many of a hundred such readers would answer correctly? The quoted item
contains its question, lettered choices and closing instruction.
We call this unchanged forecast template, rendered with the checkpoint's chat
template and without an answer stem, the \emph{production prompt}; its
\emph{original position} is the next token after the assistant turn's opening.
The \emph{answer stem} \code{"Out of 100 such readers, about\ "}, ending in a
space, is appended after that opening; the \emph{stemmed position} is
the next token after it. \code{lm-evaluation-harness} \citep{lmevalharness}
provides such a response prefix (Appendix~\ref{app:harness}).
In the downstream task, one checkpoint writes the note, reads it as the
note-only reader and forecasts; the target is that one reader's correctness,
and ``Out of 100 such readers'' is the question's wording.

Items come from RACE-H \citep{lai2017race} passages of at least 200 words, with
one question per passage. Every RACE-H sample is a prefix of the same shuffled
list. Appendices~\ref{app:sampling} and~\ref{app:consequence} specify sampling and note
budgets. The main text uses three RACE-H samples: 64 items with one fixed note
(Table~\ref{tab:crossfamily}, Sections~\ref{sec:failure} and~\ref{sec:families});
120 rows on Qwen3-4B with constructed notes, half stating the answer and half
about an unrelated passage (Section~\ref{sec:failure}); and 240 passages with
the notes each checkpoint writes itself (Table~\ref{tab:consequence}).
Appendix~\ref{app:general} also tests notes written for individual items.
The text shortens checkpoint names, for example to Phi-3 for
\code{Phi-3-mini-4k-instruct}; Table~\ref{tab:crossfamily} gives the full names.
The judge reads the same first 32 items across all 49 checkpoint--instruction
combinations; the experiment map lists the other samples and readouts.
Intervals are 95\% percentile bootstrap intervals, and a Qwen3-averaged value
is the mean of the three Qwen3 checkpoints' values.

\paragraph{Read the whole next-token distribution.}
Let $c$ be the prompt after chat-template rendering, through the start of the
assistant turn. At the scored position, $p(t\mid c)$ is the probability of the
next token $t$. We record its unrestricted argmax as well as the
\emph{in-context first-token mass}, or \emph{option mass}, $\ctxmass(c)$: the sum of probabilities of
the options' distinct first tokens when each option is encoded as a
continuation of $c$. Each token is counted once, even if several options share
it. This mass upper-bounds the total probability of the full option sequences;
Appendix~\ref{app:metric} gives the formula and its assumptions. The
\emph{answer-letter rate} is the fraction of items whose argmax is an A--D letter.

We call a median option mass below the diagnostic floor of $\floorval$
\emph{collapse}. On an item at or above the floor, at least half the next-token
probability falls on the options' first tokens; the unrestricted argmax shows
whether the most probable next token is one of them. Whether a score over
tokens the model is not about to emit ranks a target is a separate, measured
question: it does for lm-polygraph's P(True) renormalised over True and False
(Section~\ref{sec:failure}); the forecast renormalised at its original position
is indistinguishable from chance (Table~\ref{tab:consequence}). Shared first tokens cannot
distinguish the options, even at high mass. The \emph{unrestricted continuation},
the greedy output there, shows what the model writes.

\paragraph{The forecast that is evaluated downstream.}
Each checkpoint greedily writes its own note of at most 32 words. As the
note-only reader, it picks the option among
\code{" A"}, \code{" B"},
\code{" C"} and \code{" D"}
with the highest full-sequence log-probability; matching gold defines one
deterministic outcome per passage, and the forecast is evaluated as a
probability of that outcome. Appendix~\ref{app:consequence}
gives the prompts and counts of correct reader answers. On these passages and
the three Qwen3 checkpoints, the spaced and unspaced strings select the same
option on 96.25\% to 98.33\% of rows, and the unspaced pick equals the reader's
greedy answer on every row where one is parsed; scored against either, the
Qwen3-averaged gain of Section~\ref{sec:repair} is 0.1174 $[0.0382, 0.1969]$ and
0.1163 $[0.0371, 0.1956]$ (Appendix~\ref{app:reader}).
The paired comparison in Section~\ref{sec:repair} uses each full option's
log-probability $\ell_o$ to obtain a probability of reader correctness:
\begin{equation}
p_S=\frac{\sum_{o\in\optset}(\operatorname{int}(o)/100)\exp(\ell_o)}
             {\sum_{o\in\optset}\exp(\ell_o)}.
\label{eq:forecast-probability}
\end{equation}
Here $\operatorname{int}(o)$ is the option's number. Renormalisation gives a
weighted mean even when the options' total probability is small. We evaluate
ranking with AUC and squared probability error with Brier score;
Appendix~\ref{app:consequence} specifies encoding and the paired comparison.

\begin{table}[t]
\centering
\footnotesize
\setlength{\tabcolsep}{3pt}
\renewcommand{\arraystretch}{1.0}
\caption{\textbf{Experiment map.} Rows are numbered for reference; $n$ is per
checkpoint and condition unless specified otherwise. ``Seven'' refers to
Table~\ref{tab:crossfamily}; ``eight'' adds Llama-3-8B-Instruct; ``nine'' also
adds Qwen3-32B; ``ten'' adds Flan-T5-XL, Flan-T5-XXL and Flan-UL2 to the seven;
``eleven'' also adds GPT-2.}
\label{tab:expmap}
\begin{tabular}{@{}r@{\hspace{5pt}}p{.22\linewidth}p{.235\linewidth}p{.125\linewidth}p{.18\linewidth}p{.125\linewidth}@{}}
\toprule
& \raggedright What varies & \raggedright Checkpoints & \raggedright $n$ & \raggedright What is read & \raggedright Reported in  \tabularnewline
\midrule
1 & \raggedright Note states the answer / unrelated note & \raggedright Qwen3-4B & \raggedright 120 rows, 60 each & \raggedright Mass; unrestricted argmax & \raggedright \S\ref{sec:failure}, App.~\ref{app:probe120}  \tabularnewline
2 & \raggedright Checkpoint; answer stem & \raggedright Nine & \raggedright 64 & \raggedright Mass; argmax; below-floor counts & \raggedright Table~\ref{tab:crossfamily}, \S\ref{sec:families}  \tabularnewline
3 & \raggedright Delete / substitute instruction; choices; separator & \raggedright Seven & \raggedright 64 & \raggedright Letter rates and token classes & \raggedright \S\ref{sec:failure}, App.~\ref{app:instruction-experiment}  \tabularnewline
4 & \raggedright Judge and human labels & \raggedright Seven & \raggedright 32 per ending; 100 labelled by hand & \raggedright Execution labels and agreement & \raggedright \S\ref{sec:failure}, App.~\ref{app:instruction-experiment}  \tabularnewline
5 & \raggedright lm-polygraph P(True): library, renormalised, stemmed, deleted, labels in place & \raggedright Qwen3-4B, 8B, 14B; Llama-3-8B-Instruct; option mass also on Phi-3, Mistral, falcon, SmolLM2, Qwen2.5-72B-Instruct & \raggedright 500 MMLU; 500 TriviaQA & \raggedright AUROC; prediction--rejection ratio; option mass; argmax class & \raggedright \S\ref{sec:failure}, App.~\ref{app:lmpolygraph}  \tabularnewline
6 & \raggedright Leading spaces in reader options & \raggedright Qwen3-4B, 8B, 14B & \raggedright 240 passages & \raggedright Choice agreement; mass; accuracy; forecast AUC & \raggedright \S\ref{sec:protocol}, App.~\ref{app:reader}  \tabularnewline
7 & \raggedright Quoted item in the BEGIN/END fence; answer stem & \raggedright Qwen3-4B & \raggedright 64 & \raggedright Option mass and argmax & \raggedright \S\ref{sec:repair}, App.~\ref{app:ladder}  \tabularnewline
8 & \raggedright Isolated / in-context encoding & \raggedright Seven & \raggedright 64 & \raggedright Mass; token distinctness & \raggedright \S\ref{sec:repair}, App.~\ref{app:crossfamily}  \tabularnewline
9 & \raggedright Option strings; scored context & \raggedright Eleven; tokenisers only & \raggedright 64 contexts & \raggedright Distinct first tokens & \raggedright \S\ref{sec:repair}, App.~\ref{app:declared-sets}  \tabularnewline
10 & \raggedright Published option sets; scored positions & \raggedright Ten & \raggedright 64 requested & \raggedright Mass; option-token argmax & \raggedright \S\ref{sec:repair}, App.~\ref{app:harness}  \tabularnewline
11 & \raggedright G-Eval score; rating dimension & \raggedright Seven & \raggedright 100 documents; 1600 rows & \raggedright Correlation with human ratings & \raggedright \S\ref{sec:repair}, App.~\ref{app:geval-dimensions}  \tabularnewline
12 & \raggedright Answer stem; item's instruction deleted; another template; the generated number & \raggedright Qwen3-4B, 8B, 14B; Llama-3-8B-Instruct & \raggedright 240 passages & \raggedright Forecast AUC and Brier & \raggedright \S\ref{sec:intro}, \S\ref{sec:repair}, App.~\ref{app:consequence}  \tabularnewline
13 & \raggedright Notes; task placement; items; instruction wording & \raggedright Eight & \raggedright 64 & \raggedright Mass; argmax; unrestricted continuation & \raggedright App.~\ref{app:general}  \tabularnewline
\bottomrule
\end{tabular}
\end{table}

\section{Which instruction controls the scored position?}
\label{sec:failure}

\paragraph{The model answers the quoted item first.}
On Qwen3-4B's 120 constructed-note rows, the unrestricted argmax is an A--D
answer letter on 118. None has a forecast option's first token
as its argmax. The median option mass is $1.42\times10^{-15}$
(Appendix~\ref{app:probe120}). Figure~\ref{fig:failure-example} shows one
of these items: its unrestricted continuation answers the item before
forecasting. Greedy continuations of at most 16 new tokens of the production prompt with the
fixed note, classed by fixed rules on their opening, answer the item first on 62
or 63 of the 64 items on Qwen3-4B, 8B and 14B and on Phi-3; on those three
Qwen3 checkpoints a grid forecast follows within that limit on 61 to 63 of them.
Three people's labels agree with the rules on 92 of 100 sampled continuations,
97.9\% when reweighted to all 512 on eight checkpoints
(Appendix~\ref{app:general}).
This also occurs when the note cannot supply an answer to copy:
58 of the 60 rows with notes about an unrelated passage have answer-letter
argmaxes (Appendix~\ref{app:prompt}).

\paragraph{Deleting the item's instruction.}
The quoted item ends with \code{Answer with exactly one of: A, B, C, D.};
the forecast instruction comes after it. We remove or replace the item's
instruction on the same 64 items, holding the note fixed, and also test deletion
of its lettered choice list. Deletion is measured at the original ending on all
seven checkpoints. Substitution uses that ending on five; on Mistral and falcon,
it uses an appended space to supply the separator their templates omit.

At the original ending, deleting the item's instruction reduces the
answer-letter rate to 0.000 on the three Qwen3 checkpoints, to 0.016 on
SmolLM2, and from 0.469 to 0.016 on Mistral.
On these five checkpoints, the answer-letter argmax depends on that instruction.
On Phi-3, the rate falls to 0.500 and reaches
0.000 only when the choice list is removed as well. Its forecast options share
a first token on both prompts, so only the letter rate is interpretable.
Supplying the missing space with the instruction intact also reduces
Mistral's rate to 0.016; falcon's remains 0.000.
Appendix~\ref{app:instruction-experiment} gives the deletion and substitution results,
including the mass changes at the original ending, where the separator defect
remains on Mistral and falcon.

\paragraph{Interpreting generated continuations.}
Replacing the item's instruction changes the outputs, but two simple rules
fail to explain how: matching the instruction's opening word, and predicting
the class of its first response token (Appendix~\ref{app:instruction-experiment}). We instead
ask Grok 4.6 to label whether each continuation executes the replacement
instruction on the material shown, answers the item, forecasts, or does none of
these. The two instruments read the same cells: the continuation begins with
the recorded argmax on 1566 of 1568 rows. Under the translation
instruction, the opening-word rule reads 0.0000 on all seven checkpoints, while
the judge reads execution on six, at rates from 0.2188 to 0.9688; falcon reads
0.0000. A bare number, a continuation that is only a number, is counted as a
forecast. Two judges from other developers, Claude Opus 5.5
and GPT-6-astra, read translation execution on the same six and agree with
Grok's execution labels on 1549 and 1532 of the 1568 continuations. At the
original ending, all three judges label 31, 32, 31 and 32 of 32 continuations
as item answers on Qwen3-4B, 8B, 14B and Phi-3 with the item's instruction
intact, and all 32 on each Qwen3 checkpoint as forecasts with it deleted.
Three people labelled a separate sample of 100 continuations without seeing
the judge's labels; Grok agrees with their majority on execution on all 99 they
label unanimously. Table~\ref{tab:judge-agreement} adds a second labelling and
the other agreements (Appendix~\ref{app:instruction-experiment}).

A maintained implementation, lm-polygraph's default P(True) estimator
\citep{fadeeva2023lm}, reads \code{True} where the model writes another token
(Appendix~\ref{app:lmpolygraph}). On 500 MMLU items, an A--D letter is the
unrestricted argmax on 387, 496 and 500 items on Qwen3-4B, 8B and 14B without
thinking. Of these, 179, 215 and 129 are C or D, which the prompt's
\code{(A) True} and \code{(B) False} labels do not offer. Once the quoted item's
instruction is deleted, a letter is the argmax on 0, 0 and 3 items. On 500 TriviaQA items, the library's own labels
\code{(A} and \code{(B} hold the position on 463 to 500 items per Qwen3 checkpoint.
On the three Qwen3 checkpoints, \code{True} and \code{False} hold a median mass
of at most $1.2\times10^{-10}$ on the 500 MMLU items and $5.1\times10^{-8}$ on the 500 TriviaQA items.

\begingroup
\setlength{\topsep}{2pt}\setlength{\partopsep}{0pt}%
\captionsetup{font=footnotesize,skip=3pt}%
\begin{center}
\captionof{table}{\textbf{lm-polygraph P(True) readouts.} Qwen3-averaged AUROC
for the model's own answer being correct, 500 items per task;
Tables~\ref{tab:lmpolygraph-mmlu} and~\ref{tab:lmpolygraph-trivia} give each
checkpoint for Library, Renormalised, Moved and Deleted. Library: $\log P(\code{True})$ at the library's position.
Renormalised: \code{True} against \code{False} there. Labels in place:
\code{(A} against \code{(B} there. Moved: \code{A} against \code{B} after the
stem \code{The possible answer is (}. Deleted: Library without the quoted
instruction.}\label{tab:lmp-readouts}
\scriptsize
\setlength{\tabcolsep}{3pt}
\begin{tabular}{@{}lccccc@{}}
\toprule
& Library & Renormalised & Labels in place & Moved & Deleted \\
\midrule
MMLU & 0.503 [0.464, 0.540] & 0.660 [0.624, 0.694] & 0.563 [0.520, 0.606] & 0.632 [0.595, 0.669] & 0.587 [0.552, 0.623] \\
TriviaQA & 0.473 [0.448, 0.497] & 0.860 [0.836, 0.882] & 0.857 [0.833, 0.880] & 0.868 [0.844, 0.890] & 0.731 [0.704, 0.758] \\
\bottomrule
\end{tabular}
\end{center}
\endgroup
The library's P(True) confidence is indistinguishable from chance at ranking
the model's correct answers on MMLU and below chance on TriviaQA; Qwen3-4B's
library values are 0.423 $[0.369, 0.477]$ and 0.185 $[0.147, 0.226]$. After the stem, Moved's
declared set holds the position: on the three Qwen3 MMLU runs, its medians
exceed 0.999 and 398, 479 and 494 of 500 items reach the floor. Its paired gain
over the library score is 0.129 $[0.078, 0.181]$ on MMLU, and on TriviaQA it is
above all fourteen library estimators in the single-answer comparison. Moved
minus Renormalised is $-0.028$ $[-0.058, +0.003]$ on MMLU and $+0.008$
$[-0.002, +0.018]$ on TriviaQA. Of Library, Renormalised and Moved, only Moved's
declared set holds the scored position; in the forecast task the stem ranks
correctness where renormalising at the original position stays
indistinguishable from chance (Section~\ref{sec:repair}). For Labels in place,
an answer letter is usually the argmax on MMLU and a label on TriviaQA; the stem
adds 0.069 $[0.042, 0.095]$ and 0.011 $[0.001, 0.021]$ to it. On four further checkpoints, Phi-3, Mistral, falcon and
SmolLM2, \code{True} and \code{False} also hold under half the mass at the
library's position on every item of both tasks. On Qwen2.5-72B-Instruct, their
median mass there is $9.85\times10^{-7}$ on MMLU and $4.27\times10^{-6}$ on
TriviaQA. With a non-gold letter printed as the possible answer, the MMLU
argmax is the gold C or D on 102, 132 and 80 of 500 items on the Qwen3
checkpoints (Appendix~\ref{app:lmpolygraph}).

\section{How often the readout fails across checkpoints}
\label{sec:families}

Table~\ref{tab:crossfamily} compares seven checkpoints from five architecture
families, using the same 64 items and fixed note as Section~\ref{sec:failure}.

\paragraph{Six low-mass cases and one partial reproduction.}
All seven production-prompt medians meet the collapse definition. Six are at
or below $5.8\times10^{-4}$, four of them below $10^{-5}$, and every item on
these six is below the floor, and no item's unrestricted argmax is a forecast
option's first token. SmolLM2, the partial reproduction, has a median of
0.3895, with 50 of 64 items below the floor; 18 of its 64 argmaxes are an
option's first token. Read the same way, the larger \code{Qwen3-32B} has a median of $3.79\times10^{-3}$;
all 64 items are below the floor and none has a forecast option at the
argmax. The stem lifts its median to 0.99928
(Appendix~\ref{app:general}). With Llama-3-8B-Instruct (median
$2.565\times10^{-4}$, stemmed 0.99352), all nine checkpoints read with this prompt
start below the floor and the stem lifts all nine above it.
Table~\ref{tab:crossfamily} gives each per-item minimum beside its median.
The options' mass and the argmax check part on 13 of the 1024 item readings of
Table~\ref{tab:crossfamily} and Qwen3-32B, where the argmax is an option but the
options hold under half the mass or the reverse (Appendix~\ref{app:crossfamily}).

\paragraph{Answer letters on five checkpoints, missing spaces on two.}
Answer letters occupy the scored position on most items for the three Qwen3
checkpoints, Phi-3 and SmolLM2 (Table~\ref{tab:crossfamily}). Mistral's most
frequent next token is instead the bare space marker \sq{}, on 30 of 64
items. Answer letters take another 30, \sq{D} alone 27, a letter rate of
0.469. Falcon produces the space token \gq{} on 35 of 64 and the newline
\cq{} on 29, with a letter rate of 0.000. Both rendered prompts omit final
whitespace. Supplying the space puts a digit at the scored position and
reduces Mistral's letter rate to 1 of 64; falcon's remains 0.000. On these two,
at most one item in 64 keeps an answer letter once the separator is supplied
(Appendix~\ref{app:repair-details}).

Phi-3 has a second problem: its options share a first token even when encoded
in context. Its median of $5.79\times10^{-4}$ therefore measures option mass
without establishing that the options are separable. Its answer-letter rate
of 0.984 independently establishes the failure.

\begin{table}[h]
\centering
\renewcommand{\arraystretch}{0.95}%
\small
\caption{Seven checkpoints on the same 64 frozen items, with one note held fixed:
\code{Landlord only gives when told to take, not give.} It names no answer
letter or option. Each checkpoint uses its own chat template, with
\code{enable\_thinking=False} where the template reads it; forward passes
use float16 and log-softmax uses float32. Appendix~\ref{app:crossfamily}
gives the note source. $\ctxmass$ (Appendix~\ref{app:metric}) is the median over items, per-item
minimum in parentheses; \emph{Letter} is the fraction of
items whose unrestricted argmax is an A--D answer letter; \emph{Below} counts
items under \floorval. }
\label{tab:crossfamily}
\begin{tabular}{@{}lrrrrr@{}}
\toprule
& \multicolumn{3}{c}{Production prompt} & \multicolumn{2}{c}{With answer stem} \\
\cmidrule(lr){2-4}\cmidrule(lr){5-6}
Checkpoint & $\ctxmass$ (min) & Letter & Below & $\ctxmass$ (min) & Below \\
\midrule
Qwen3-4B-Instruct-2507 & \num{2.71e-15} (\num{7.63e-19}) & 0.984 & 64/64 & 0.99940 (0.17702) & 2/64 \\
Qwen3-8B & \num{8.14e-14} (\num{6.97e-17}) & 0.953 & 64/64 & 0.99951 (0.92296) & 0/64 \\
Qwen3-14B & \num{1.37e-08} (\num{2.96e-14}) & 0.969 & 64/64 & 0.99999 (0.99966) & 0/64 \\
Phi-3-mini-4k-instruct & \num{5.79e-04} (\num{9.07e-05}) & 0.984 & 64/64 & 0.93344 (0.84818) & 0/64 \\
Mistral-7B-Instruct-v0.3 & \num{3.49e-06} (\num{3.67e-08}) & 0.469 & 64/64 & 0.99050 (0.00216) & 7/64 \\
falcon-7b-instruct & \num{1.05e-04} (\num{5.70e-05}) & 0.000 & 64/64 & 0.53214 (0.33557) & 5/64 \\
SmolLM2-1.7B-Instruct & 0.3895 (0.0352) & 0.719 & 50/64 & 0.88834 (0.80527) & 0/64 \\
\bottomrule
\end{tabular}
\end{table}

\paragraph{The answer stem's scope.}
The stem lifts all seven medians above the floor on Table~\ref{tab:crossfamily}'s
prompt. Falcon's argmax is a forecast option on every item, with winning-option probabilities
between 0.122 and 0.315 and roughly 47\% of mass off the grid. Mistral's worst
item stays more than two orders of magnitude below the floor. The table gives
per-item minima and below-floor counts.

Appendix~\ref{app:general} adds Llama-3-8B-Instruct. With per-item
notes, a system-turn instruction, MMLU-Pro items \citep{wang2024mmlupro} or a sentence assigning the
item's instruction to the readers, medians stay below the floor on 8, 6, 8
and 8 of eight checkpoints; an item answer letter is the unrestricted argmax on
at least half the items, which we call \emph{letter pull}, on 5, 5, 3 and 4
checkpoints, respectively
(Table~\ref{tab:mechanisms}). Llama-3-8B-Instruct's article \code{A}, which opens prose such
as \code{A clever question!}, is not counted as an answer letter. Adding a second sentence leaves
five medians below the floor.

Two whole forecast templates, T1 and T2, written by a language model from a
task description without knowledge of the failure, quote the same item block
and end in an explicit output contract. Both remove letter pull on every checkpoint.
Each leaves three medians below the floor, none of them from letter pull:
Mistral's and falcon's, whose templates omit the final separator, with
Phi-3's under T1, whose options share a first token, and SmolLM2's under T2,
which puts an option at the argmax on 56 of 64 items with no item reaching the
floor (median 0.31081). With their own answer stems, \code{Estimate:\ } for T1 and
\code{Predicted number correct:\ } for T2, all eight medians clear the floor under T1 and seven under T2. Falcon's stemmed medians are 0.45421 with MMLU-Pro
and 0.33207 under T2 (Tables~\ref{tab:general} and~\ref{tab:mechanisms}).

\section{Moving the scored position and checking the readout}
\label{sec:repair}

\paragraph{The downstream forecast.}
The practical question is whether the scored forecast tracks the correctness of
its note-only reader. Across 240 passages on each of the three Qwen3 checkpoints and
Llama-3-8B-Instruct, Table~\ref{tab:consequence} compares the original and
stemmed positions. At the original position, the Qwen3-averaged AUC interval
includes chance, and every Qwen3 passage is below the mass floor
(Appendix~\ref{app:consequence}).
The declared set and renormalisation are fixed and the stem moves the scored
position, so the paired Qwen3-averaged gain, 0.1223 $[0.0449, 0.1993]$, compares the
two positions under one readout; Qwen3-14B's own difference interval includes
zero. Both readouts use the same model, precision, GPU,
note, item and option first-token IDs (Appendix~\ref{app:consequence}).
On Llama-3-8B-Instruct, AUC rises from 0.5231 to 0.6044, a paired
difference of 0.0813 $[0.0046, 0.1574]$. The stemmed Brier scores, 0.2544 for
the Qwen3 average and 0.2522 for Llama-3-8B-Instruct, remain worse than the
0.2500 of a constant 0.5 forecast; Table~\ref{tab:constant-baselines} adds
constants fitted to the evaluation rows.
In Table~\ref{tab:consequence-readouts}'s second reading of these passages, on
other hardware, a short stem, \code{Number: }, gives a Qwen3-averaged AUC of
0.6145 against the task stem's 0.6156, a difference of $-0.0011$
$[-0.0213, 0.0191]$; for \code{Forecast: } the difference is $-0.0084$
$[-0.0328, 0.0162]$ (Appendix~\ref{app:consequence}).
Deleting the item's instruction instead of adding the stem ranks reader
correctness worse: averaged over the four checkpoints, its AUC is lower than the
stem's by 0.0489 $[0.0191, 0.0791]$, although on the Qwen3 checkpoints the
deletion also returns the options' median mass above 0.9998.
On the same four-checkpoint average, template T1 (Section~\ref{sec:families})
comes within 0.0063 $[-0.0160, 0.0291]$ of the stem's AUC with its own answer
stem; without it, the gap is 0.0250 $[-0.0019, 0.0518]$.
Reading the number off the model's own unrestricted continuation of the
production prompt ranks worse than the stem too, by 0.0772 $[0.0355, 0.1169]$ on that
average, with a continuation lacking a grid number scored 0.5; those forecasts are
nearly constant, 70 on 232 of Qwen3-14B's 240 passages (Appendix~\ref{app:consequence}).

\begin{table}[!ht]
\centering\small
\caption{The forecast scored against the correctness it predicts, at the original
and stemmed positions, on the same 240 passages
at each of the three Qwen3 checkpoints and Meta-Llama-3-8B-Instruct. A Qwen3-averaged row is the mean of the three
checkpoints' values; Llama-3-8B-Instruct is separate. Intervals are 95\% percentile intervals of
a paired cluster bootstrap over passages, $B=20{,}000$, both readouts resampled on the same replicates.}
\label{tab:consequence}
\setlength{\tabcolsep}{5pt}
\begin{tabular}{@{}lrrr@{}}
\toprule
& Original position & With answer stem & Difference \\
\midrule
\multicolumn{3}{@{}l}{$\mathrm{AUC}$} & Gain, stemmed $-$ original \\
Qwen3-averaged & 0.4929 [0.4390, 0.5458] & 0.6152 [0.5602, 0.6686] & 0.1223 [0.0449, 0.1993] \\
\quad Qwen3-4B & 0.4958 & 0.6283 & 0.1325 [0.0168, 0.2459] \\
\quad Qwen3-8B & 0.4864 & 0.6435 & 0.1570 [0.0431, 0.2685] \\
\quad Qwen3-14B & 0.4964 & 0.5739 & 0.0775 [$-$0.0199, 0.1744] \\
Llama-3-8B-Instruct & 0.5231 [0.4486, 0.5961] & 0.6044 [0.5312, 0.6764] & 0.0813 [0.0046, 0.1574] \\
\midrule
\multicolumn{3}{@{}l}{Brier} & Reduction, original $-$ stemmed \\
Qwen3-averaged & 0.4749 & 0.2544 & 0.2205 [0.1844, 0.2559] \\
\quad Qwen3-4B & 0.4487 & 0.2693 & 0.1794 [0.1399, 0.2190] \\
\quad Qwen3-8B & 0.5350 & 0.2463 & 0.2887 [0.2400, 0.3353] \\
\quad Qwen3-14B & 0.4409 & 0.2475 & 0.1933 [0.1439, 0.2414] \\
Llama-3-8B-Instruct & 0.2640 & 0.2522 & 0.0118 [$-$0.0152, 0.0387] \\
\bottomrule
\end{tabular}
\end{table}

\paragraph{What the answer stem changes.}
For Table~\ref{tab:crossfamily}'s production prompt, the stem survives
tokenisation and decoding at the end of the context on all seven checkpoints
and lifts their medians above the floor. On five checkpoints the context ends
in whitespace both before and after adding the stem, yet the medians rise from
at most 0.390 to at least 0.888. Trailing whitespace alone therefore does not
explain the increase. On Phi-3, the stem also restores distinct first
tokens by ending the stemmed context in a space rather than a newline.
Our \emph{fence} instead wraps the quoted item in BEGIN/END markers and adds a
request not to answer it, adapted from the DIRECT prompt of
\citet[Section~5.1]{hwang2025distraction}, tested there on
Llama-3.1-70B-Instruct translation. It leaves 43 of 64 Qwen3-4B items
below the floor, compared with two for the stem
(Appendix~\ref{app:ladder}).

\paragraph{First-token scoring needs distinct first tokens.}
Tokenising each option on its own gives every option the same initial
word-boundary token on the two sentencepiece checkpoints among the seven, where
the resulting mass measures that token's probability. Encoding each option as
a continuation of the rendered prompt removes this isolated-encoding error,
but the prompt's ending can still produce a shared first token. We cross three
prompt forms (original, fenced item, original with the answer stem) with three
item endings (original, instruction deleted, summarisation instruction).
On Phi-3, the six combinations without the stem share a first token and
the three stemmed ones have distinct first tokens; the other six checkpoints
have distinct first tokens on all nine (Appendix~\ref{app:crossfamily}).
Moving the scored position requires repeating this check.

A large mass cannot substitute for distinctness. G-Eval's prompt
\citep{liu2023geval} ends at
\mbox{\code{- Coherence:}} with no separator; adding a space to each G-Eval option raises the mass on Qwen3-4B from $1.26\times10^{-5}$ to 0.99449,
but all five options then share the space token on all seven checkpoints.
The prefix trades the separator defect for the shared-first-token defect.
Which defect a declared set meets is decided by how the vocabulary tokenises
it, not by the strings alone (Appendix~\ref{app:declared-sets}).

\paragraph{Agreement with ratings.}
Appendix~\ref{app:harness} gives the four specifications cell by cell
\citep{liu2023geval,zhuang2023beyond,dubois2024length,kadavath2022language}.
On G-Eval's space-prefixed options, first-token scores are constant on all
seven checkpoints, so their rank correlations are undefined. Supplying the
separator in the context and scoring full option sequences makes the
comparison defined, with correlation changes in both directions
(Appendix~\ref{app:geval-dimensions}).

\paragraph{Checks before using a constrained-option score.}
Each check addresses a failure measured here.
\begin{enumerate}
\setlength{\itemsep}{0pt}
\setlength{\parskip}{0pt}
\item Measure the options' in-context first-token mass at the scored position.
A median below 0.5 means that on at least half the items most next-token
probability lies outside the options' first tokens (Section~\ref{sec:protocol}).
\item Check the unrestricted argmax at the scored position. An answer letter
where a forecast is requested flags a possible answer to the quoted item
(Section~\ref{sec:failure}).
\item Inspect the unrestricted continuation from the scored position. It shows
whether the model answers the quoted item before forecasting, and so what role
a flagged token plays (Figure~\ref{fig:failure-example}).
\item Verify that the options' first tokens are distinct when encoded as
continuations of the rendered prompt. Shared first tokens cannot distinguish the
options, even at high mass (Section~\ref{sec:repair}).
\item Inspect the rendered prompt's final characters and test a missing
separator. A chat template that omits it can put whitespace at the scored
position (Section~\ref{sec:families}).
\item Test an answer stem, repeat these checks after it and recheck the
downstream target. On Table~\ref{tab:crossfamily}'s prompt the stem lifts all
seven median masses above the floor; the Brier score delimits what that lift
establishes (Section~\ref{sec:repair}).
\end{enumerate}

\section{Limitations and conclusion}
\label{sec:limitations}

The main mass and token-class comparisons use RACE-H, one forecast template,
64 items per checkpoint and one fixed note. Table~\ref{tab:expmap} gives the other sample sizes;
Appendix~\ref{app:general} varies the note, dataset, turn and template and adds
Llama-3-8B-Instruct. The answer-stem result on nine checkpoints is
scoped to Table~\ref{tab:crossfamily}'s prompt.
The Qwen3 checkpoints hold architecture and tokeniser fixed; from 4B to 32B,
each keeps its production-prompt median below the floor, and the answer stem
lifts each above it. Scale varies only within Qwen3, so these comparisons establish
no size trend across families. SmolLM2 tests whether the failure is specific
to Qwen3, not whether it is specific to small models.

Two observations remain unexplained. Deleting the imperative reduces Mistral's
median option mass at the original ending from $3.49\times10^{-6}$ to
$1.77\times10^{-6}$, with the separator still omitted. The token-class
rules also fail to predict which class replaces the answer letters.
The method requires log-probabilities for chosen tokens at a chosen position,
so these measurements do not address API-only models.

A constrained-option score reports a number even when the model is about to
emit something else. Self-evaluation and judge scorers have this configuration
when the quoted material carries its own answer instruction and the scorer
reads a different declared set (Appendix~\ref{app:lmpolygraph}). Inspecting the unrestricted continuation identifies what
is being answered; checking the tokenised options and rendered ending identifies
what is being scored. The paired downstream comparison then tests whether that
score tracks the intended target.

\subsubsection*{Reproducibility Statement}

Section~\ref{sec:protocol} and the appendices give the readouts, prompts, sampling and statistical comparisons. The supplementary material provides the per-item readings behind Table~\ref{tab:crossfamily} and its Qwen3-32B extension, the forecast comparison of Table~\ref{tab:consequence} and Appendix~\ref{app:consequence}, and the lm-polygraph comparisons of Section~\ref{sec:failure} and Appendix~\ref{app:lmpolygraph}, together with the human and judge labels behind the execution-label agreement reported in Section~\ref{sec:failure}; its \code{README.md} names the fields and the resampling convention for recomputing each statistic. Appendix~\ref{app:snapshots} pins the checkpoints of Table~\ref{tab:crossfamily} and the checks that confirm them; Appendices~\ref{app:consequence} and~\ref{app:general} name the snapshots of the checkpoints they add; the cited \code{lm-evaluation-harness} counts and source locations use the revision pinned in Appendix~\ref{app:snapshots}.

\subsubsection*{Use of AI}

In this work, we used generative AI tools to generate synthetic data sets, to
search for, identify and summarise relevant literature, and to edit and polish
the text of this paper for readability. We have not used generative AI tools to help develop theoretical
models or conceptual frameworks, to interpret results, to propose or refine
hypotheses, to design or give feedback on research methodology or experiments,
to implement methods, to clean and reformat datasets, or to support qualitative
and thematic data analysis. Formulating
mathematical claims, providing critical ingredients for proving mathematical
claims, assisting in the writing of proofs, and translation are not applicable
to this work. Additionally, we used generative AI tools for the creation of
artifacts.

Separately from assistance with the research, language models are components
of the experiments, reported with their results: three judge models label
whether each continuation executes an instruction (Section~\ref{sec:failure},
Appendix~\ref{app:instruction-experiment}), with their labels checked against two independent three-person labellings,
and a language model wrote the two forecast templates T1 and T2 from a task
description, without knowledge of the failure, as test inputs
(Appendix~\ref{app:general}). A language model also wrote the synthetic test inputs
with known answers that check the labels: the twelve validation cases of
Appendix~\ref{app:instruction-experiment} and the ten practice records of Appendix~\ref{app:general}.

We have reviewed all AI-assisted work. The papers we cite were read in full, and
a citation was checked at the place in the cited paper that supports the claim
it carries; the authors edited the text and verified it against the results
it reports.

We take responsibility for the final content of this work, including text,
claims or artifacts produced with the aid of generative AI.

\bibliographystyle{iclr2027_conference}
\bibliography{references}

\appendix
\section{Definitions and sampling}
\label{app:metric}

The original ending is the unmodified rendered prompt;
the space-appended ending adds one space to it.

Section~\ref{sec:protocol} defines $\ctxmass$ in words; this is the closed form.
Each option $o \in \optset$ is tokenised \emph{as a continuation of} the rendered
context $c$: if $\mathrm{tok}(c)$ prefixes $\mathrm{tok}(c \| o)$, the option's
tokens are the suffix and $t_1(o \mid c)$ its first. If this prefix condition
fails for any option, the implementation marks the boundary as not preserved
and records no in-context mass (\code{None}), rather than zero. When the
condition holds for every option,
\begin{equation}
\ctxmass(c) \;=\; \sum_{t \,\in\, \{\, t_1(o \mid c) \;:\; o \in \optset \,\}} p(t \mid c).
\label{eq:mass}
\end{equation}
When no option's token sequence is a prefix of another's, the full options'
probabilities sum to at most $\ctxmass(c)$. The sum counts each token once and measures separability only when the
options' first tokens are distinct. On Qwen3-4B's 16 ladder items with the
fence, the stem and space-prefixed options, a list-based sum reports a median
of 4.9850 where the set sum is 0.997, and equal token lengths do not detect
the duplicate first tokens. Scoring after a supplied shared prefix measures a different
conditional position, giving Phi-3 a median of 0.0368 instead of 0.9334 on the
64-item stemmed prompt; Appendix~\ref{app:isolated-encoding} gives the 16-item
reading, 0.0392 against 0.9331.

\subsection{Sampling}
\label{app:sampling}
RACE-H is read from Hugging Face \code{ehovy/race}, configuration \code{high},
at revision \code{2fec9fd81f1dc971569a9b729c43f2f0e6436637}.
We read \code{test}, \code{validation}, then \code{train} in stored row order,
grouping rows by the MD5 of the UTF-8 passage text, keeping the first
occurrence's position and accumulating its questions in encounter order.
We keep the 19{,}566 passages of at least 200 words, shuffle them with
\code{random.Random(0)}, and choose one question for each of the first
3{,}052 passages with the same generator. Every RACE-H sample in this paper
is a prefix of this list, including the 64 items per checkpoint and the
judge's first 32, shared by all 49 checkpoint--ending combinations.

\subsection{Reproduction snapshots}
\label{app:snapshots}
Table~\ref{tab:crossfamily} reads \code{Qwen/Qwen3-4B-Instruct-2507} at
\code{cdbee75}, \code{Qwen/Qwen3-8B} at \code{b968826}, \code{Qwen/Qwen3-14B}
at \code{40c0698}, \code{microsoft/Phi-3-mini-4k-instruct} at \code{f39ac1d},
\code{mistralai/Mistral-7B-Instruct-v0.3} at \code{c170c70},
\code{tiiuae/falcon-7b-instruct} at \code{8782b5c} and
\code{HuggingFaceTB/SmolLM2-1.7B-Instruct} at \code{31b70e2}. Every weight and
tokenizer file the runs read matches the Hub's record for its revision.
Loaded at these revisions, Qwen3-8B, Qwen3-14B, Mistral and falcon reproduce
their rows item for item on the GPU model they were read on; Qwen3-4B,
Phi-3-mini and SmolLM2 were read on a second GPU model, and their re-read
differs as Appendix~\ref{app:general} describes for C0.
All cited lm-evaluation-harness source locations and task counts use commit
\code{d6de81643928d653435c431bae19945d41d32520}.

\section{The prompt, rendered}
\label{app:prompt}

This is the complete forecast prompt for \code{high6240.txt} on
\code{Qwen/Qwen3-4B-Instruct-2507}, the example in Figure~\ref{fig:failure-example}.
Its last 60 characters, its 178 tokens and its last token id
match the run that scored it.
The constructed note contains the passage's first two sentences and its gold
answer. The checkpoint's chat template opens the assistant turn below.

\begin{Verbatim}[fontsize=\small,breaklines=true,breakanywhere=true,
                 breaksymbolleft={\tiny\ensuremath{\hookrightarrow}},frame=single,
                 samepage=true]
<|im_start|>user
Note:
A woman from Japan was telling a friend about her trip to the United States. The woman had visited major businesses and investment companies in New York City and Chicago. The correct answer is A: her English was poor.

A future reader will be shown ONLY the note above, not the source text, and will then be asked:
After the woman visited the United States she might feel that   _  .
A: her English was poor
B: it's easy to master English
C: it's difficult to make money
D: people there weren't very friendly
Answer with exactly one of: A, B, C, D.

Out of 100 such readers, how many answer correctly? Reply with exactly one of: 10, 30, 50, 70, 90.<|im_end|>
<|im_start|>assistant
\end{Verbatim}

\noindent
The rendered string has 704 characters; its SHA256, computed over UTF-8
with no added trailing newline, is \code{7db81d9513ef11e6c7bd8f73ad7a8459cb4b04dac22d987419f16db9dc7246d9}.
Arrows mark typeset line breaks. The string ends with the newline after
\code{assistant}; the forecast instruction follows the item's letter instruction.

\paragraph{The scored position.}
The readout scores the next token after zero-based index 177,
the newline (id \code{198}) closing the assistant opening.
The five options are \code{10}, \code{30}, \code{50}, \code{70}, \code{90};
each has two tokens with distinct first-token ids \code{16}, \code{18}, \code{20}, \code{22}, \code{24}.

On the 60 rows of the 120-row probe (Appendix~\ref{app:probe120})
whose notes concern an unrelated passage and so state no answer, an A--D letter
is still the unrestricted argmax on 58. The forecast argmax and the
reader task's argmax are the same token on 59 of the 60 rows
whose notes state the answer, against 38 of the 60
whose notes do not.

\paragraph{The scored position with the stem.}
Appending \code{"Out of 100 such readers, about\ "} to the rendered string, after the assistant
opening, adds 11 tokens; scoring follows index
188, the trailing space (id \code{220}).
On this item, the argmax is \texttt{\small '5'} at 0.98907 and the declared set
holds 0.9999999 of the mass; the unrestricted continuation is \code{"50 answer correctly.\textbackslash{}n\textbackslash{}n50"}.

\section{Supporting measurements}
\label{app:supporting}

\subsection{Cross-family readings}
\label{app:crossfamily}

\paragraph{The fixed note.}
Table~\ref{tab:crossfamily} uses Qwen3-4B's note for item 58
(\code{high2565.txt}), with its quotation marks removed. Its full text is
printed in the table caption. With the answer stem, falcon's argmax is
\code{'90'} on 49 of 64 items and \code{'10'} on the other 15; its winning
probabilities range from 0.1225 to 0.3147, with median 0.2133.

\paragraph{Option mass and the argmax check.}
On Table~\ref{tab:crossfamily}'s two prompts and the same prompts for
Qwen3-32B, 1024 item readings in all, the options' mass and the argmax check
disagree on 13. SmolLM2's production prompt puts an option at the argmax on 6
items whose options hold under half the mass and an answer letter there on 2
items whose options hold more than half; falcon's stemmed prompt puts an option
at the argmax on 5 items under half. On the other 14 checkpoint and prompt
pairs the two agree on every item.

The nine combinations cross three prompts (production, fenced, and production
with the answer stem) with three quoted-item endings (the letter instruction,
its deletion, and its replacement by \code{Summarise in one sentence.}).
Across these combinations, Phi-3's six unstemmed
combinations share a first token and its three stemmed combinations have
distinct first tokens; the other six checkpoints have distinct first tokens
on all nine. Isolated and in-context masses coincide bitwise on all nine
combinations for the five checkpoints with distinct isolated first tokens.
The two sentencepiece cases are compared in Appendix~\ref{app:isolated-encoding}.

\subsection{Details of answer-stem and boundary checks}
\label{app:repair-details}

The boundary rule's two exposure classes require opposite values of whether
the rendered context ends in whitespace (Table~\ref{tab:boundary-classes}). Exposure
is checked against in-context token distinctness and the unrestricted argmax.
The stem restores distinct first tokens on Phi-3 and supplies the missing
separator on Mistral and falcon; it leaves the whitespace condition unchanged
on five checkpoints whose medians rise from at most 0.390 to at least 0.888.
Qwen3-4B's unstemmed forecast has neither boundary defect and still begins with
an answer letter on 0.984 of items. These checks depend on the rendered context,
so the variant has to be named on every reading.

\begin{table}[h]
\centering\small
\caption{Boundary exposure classes for the measured checkpoints and tokeniser
versions. The rule separates endings with and without whitespace; the text
reports which defects occur.}
\label{tab:boundary-classes}
\setlength{\tabcolsep}{5pt}
\begin{tabular}{@{}llll@{}}
\toprule
Prompt ends with & Vocabulary & Consequence & Checkpoints \\
\midrule
Newline & sentencepiece & marker restored, mass not a reading & Phi-3-mini \\
Newline & byte-level & neither problem & Qwen3 $\times3$, SmolLM2 \\
\code{[/INST]}, \code{Assistant:} & either & separator omitted by template & Mistral, falcon \\
\bottomrule
\end{tabular}
\end{table}

\paragraph{The paired separator controls.}
Two checkpoints have the omitted-separator defect. Neither has an answer letter as its modal token across the items of any reading in
Sections~\ref{sec:failure} to~\ref{sec:repair}; in Appendix~\ref{app:general}, under the
system-turn condition S1, Mistral has a letter at the argmax on 57 of 64 items
(Table~\ref{tab:general}), with \code{A} the modal token on 18.
\code{Mistral-7B-Instruct-v0.3} has an answer-letter rate of 0.469 at the
original ending, on 30 of 64 items, histogram 30 whitespace tokens, 30 answer
letters and 4 prose words. Deleting the item's instruction reduces the rate to
0.016, leaving
63 whitespace argmaxes and one answer letter. Supplying the separator the
template omits, with the instruction kept, takes the rate to 1 of 64, with 63
digits and 1 answer letter: 29 of the 30 letter items disappear. Both
interventions remove all but one of Mistral's letters: at the original ending
they depend on the instruction, and supplying the separator removes them even
with the instruction intact. With the separator supplied, 56 of the digits are
grid-option first tokens and seven are \code{0}.
Falcon shows the same defect with 0 of 64 either way, which is why the pair is needed to tell the two apart;
supplying the separator puts a grid option's first token at the argmax on 63
of 64 falcon items. For substitution, both checkpoints use the space-appended ending;
the deletion comparison above uses the original ending.

\subsection{Scored-position probe on 120 rows (Section~\ref{sec:failure})}
\label{app:probe120}

The 120 RACE-H rows use 60 notes containing the passage's opening text and
gold answer, and 60 notes about an unrelated passage. An A--D answer letter
is the unrestricted argmax on 118 of these rows,
and no forecast option is the argmax. The median option mass is
$1.42\times10^{-15}$; its extremes are $1.73\times10^{-19}$ and
$4.64\times10^{-6}$. The two letter exceptions are \code{high1845.txt},
whose argmax is \code{'None'} at 0.868, and \code{high14114.txt}, whose
argmax is \code{'0'} at 0.99998.

\subsection{Prompt ladder}
\label{app:ladder}

The request not to answer the quoted item adapts the DIRECT prompt of
\citet[Section~5.1, Table~4]{hwang2025distraction}, tested
on Llama-3.1-70B-Instruct for translation; the BEGIN/END markers are ours.
On Qwen3-4B's 64 fixed-note items,
quoting the item gives a median mass of $2.71\times10^{-15}$ and 0 of 64
option argmaxes; fencing gives $3.74\times10^{-3}$ and 21 of 64, with mass
from $9.10\times10^{-9}$ to 0.99995; the stem gives 0.99940 and 62 of 64.
Below-floor counts are 43 with the fence and two with the stem.

\subsection{Two misleading readings from isolated encoding}
\label{app:isolated-encoding}

The columns below compare isolated, divergent (scored after the options'
shared prefix; Appendix~\ref{app:metric}) and in-context readings on the same
forward passes, for the two cases below.

For the first, on \code{Phi-3-mini-4k-instruct}'s prompt with the answer stem, the isolated
column reads $1.27\times10^{-8}$ and the divergent companion 0.0392 with
0 of 16 argmax hits, while Eq.~\ref{eq:mass} on those same 16-item
passes gives 0.9331, minimum 0.8977, and 16 of 16 hits. The argmax there is
\code{'7'} in 9 items, \code{'3'} in 6 and \code{'5'} in 1: the model emits a grid digit while
the isolated estimator scores a boundary marker after a space.
For the second, \code{Mistral-7B-Instruct-v0.3} reads 0.4545 in isolation,
which appears to be substantial option mass but is $p(\sptok)$.
Eq.~\ref{eq:mass} gives $3.93\times10^{-6}$ with 0 of 16 hits on the original ending,
where the template omits the separator.

\begin{table}[h]
\centering
\small
\caption{Isolated and divergent columns beside the in-context reading, all
from the same 16-item forward passes.
Table~\ref{tab:crossfamily} gives the 64-item medians.
A separate 16-item reading without the in-context column has identical
isolated and divergent values to every digit.}
\label{tab:isolated-encoding}
\begin{tabular}{@{}lllrrr@{}}
\toprule
& & & \multicolumn{2}{c}{Alternative columns} & Eq.~\ref{eq:mass} \\
\cmidrule(lr){4-5}\cmidrule(lr){6-6}
Reading & Checkpoint & Prompt & isolated & divergent & in-context \\
\midrule
Stem repairs the readout & Phi-3-mini-4k-instruct & stemmed & \num{1.27e-08} & 0.0392 & \textbf{0.9331} \\
Original-ending option mass & Mistral-7B-Instruct-v0.3 & unstemmed & 0.4545 & 0.9935 & \textbf{\num{3.93e-06}} \\
\bottomrule
\end{tabular}
\end{table}

\subsection{Declared sets and vocabulary}
\label{app:declared-sets}

Across three declared sets and 11 checkpoints using seven vocabularies,
\code{\{' (A)', ' (B)'\}} shares a first token on all 11 and
\code{\{' no', ' yes'\}} stays distinct on all 11. The spaced grid
\code{\{' 1',\ldots,' 5'\}} shares a first token on the seven main
checkpoints and stays distinct on GPT-2 and the three Flan checkpoints,
whose vocabularies merge the separator with the option's first characters.
A rule that predicts distinct first tokens whenever the characters after the
separator differ therefore gives seven false clearances and no false alarms
across these 33 cells. The Qwen3 models share their base vocabulary, and
Flan-T5-XL, Flan-T5-XXL and Flan-UL2 share one SentencePiece model.

\subsection{The reader task with spaced and unspaced options}
\label{app:reader}

The spaced option strings' median mass is $5.21\times10^{-14}$ where the
note contains the answer and $9.67\times10^{-12}$ where it does not. The unspaced
strings take the median to 1.000 and the argmax-on-option rate to 1.00 where the
note carries the answer and 0.967 where it does not, while
96.25\% of picks are unchanged and overall accuracy moves from 0.6958 to 0.7208.
That is the same encoding defect as Section~\ref{sec:repair}, in a third place.
The spaced strings' two group accuracies are 55.8 points apart on a
four-way item, read out of the tail with the argmax on an option in none of the
240 rows.

\begin{table}[h]
\centering
\small
\caption{The reader task with spaced and unspaced options, same 240 rows, same run. Mass is the median
option first-token mass by group; accuracy is against the passage's gold answer.
The spaced and unspaced strings select the same option in 96.25\% of rows.}
\label{tab:reader}
\begin{tabular}{@{}lrrrrrr@{}}
\toprule
& \multicolumn{2}{c}{Mass (median)} & \multicolumn{2}{c}{Argmax on option}
& \multicolumn{2}{c}{Accuracy} \\
\cmidrule(lr){2-3}\cmidrule(lr){4-5}\cmidrule(lr){6-7}
Option strings & gold in note & unrelated & gold in note & unrelated & gold in note & unrelated \\
\midrule
\code{" A"}--\code{" D"} & \num{5.21e-14} & \num{9.67e-12} & 0.00 & 0.00 & 0.975 & 0.4167 \\
\code{"A"}--\code{"D"} & 1.000 & 0.99999976 & 1.00 & 0.967 & 0.9833 & 0.4583 \\
\bottomrule
\end{tabular}
\end{table}

\paragraph{On the passages of Table~\ref{tab:consequence}.}
Each Qwen3 checkpoint's note-only reader is read again on its own notes: with
the spaced strings, which reproduces every stored pick; with the unspaced
strings; and by a greedy continuation of at most 16 tokens, taking its first
standalone letter. The spaced and unspaced picks agree on 232, 231 and 236 of
240 rows on Qwen3-4B, Qwen3-8B and Qwen3-14B, and the unspaced pick equals the
generated letter on every row whose continuation names one (240, 239 and 240
rows). Scored against the unspaced reader, the forecast's AUC moves from
0.4983, 0.4884 and 0.5019 at the original position to 0.6190, 0.6419 and
0.5799 with the stem, and the Qwen3-averaged AUC from 0.4962 $[0.4421, 0.5494]$
to 0.6136 $[0.5568, 0.6688]$, a paired gain of 0.1174 $[0.0382, 0.1969]$.
Against the generated answers, the averaged AUC moves from 0.4968 to 0.6132, a
gain of 0.1163 $[0.0371, 0.1956]$. At both positions the Brier scores move by
less than 0.01. Intervals use Table~\ref{tab:consequence}'s passage bootstrap.

\section{The instruction experiment}
\label{app:instruction-experiment}

The instruction comparison uses 64 items per checkpoint with the note held
fixed. The token-class readings and the judge's rows use the original ending on
five checkpoints and an appended space on Mistral and falcon; the deletion
comparisons of mass and of Section~\ref{sec:failure}'s letter rates use the
original ending on all seven (Appendix~\ref{app:repair-details}). The six
substitutes for the item's own \code{Answer with exactly one of: A, B, C, D.}
are the empty ending, \code{Summarise in one sentence.},
\code{Reply with a single integer.}, \code{List three keywords from the passage.},
\code{Translate the passage into French.} and
\code{Count the sentences in the passage.} The list, translation and counting
instructions name \code{the passage}; the prompt gives the model the note and
quoted item.

\paragraph{Deleting the instruction and the choice list.}
Deleting only the lettered choice list leaves answer-letter rates of 0.891,
0.781, 0.672 and 0.719 on Qwen3-4B, Qwen3-8B, Qwen3-14B and Phi-3,
respectively; SmolLM2 rises from 0.719 to 0.828. With the instruction already
deleted, Phi-3 reads 0.500 with the choices and 0.000 without them; the other
six read at most 0.016 either way. With the original rendered ending, deleting
the multiple-choice imperative raises the median mass by factors of
$3.7\times10^{14}$, $1.2\times10^{13}$ and $7.3\times10^{7}$ on the three Qwen3
checkpoints in Table~\ref{tab:crossfamily} order, 1.49 on SmolLM2 and 1.39 on
falcon; it falls by a factor of 0.507 on Mistral. The last two readings are at
the omitted separator, where deleting the imperative leaves the template
defect. Deletion raises the median above the floor on four checkpoints: the
three Qwen3 checkpoints and SmolLM2, which rises from 0.3895 to 0.58023; falcon
begins at $1.05\times10^{-4}$ and stays below the floor. Phi-3 has shared first
tokens both with and without the imperative, so only its letter rate is
interpretable.

\paragraph{Token-level rules.}
The opening-word rule asks whether the argmax is the substituted instruction's
own opening word; over the list, translation and count instructions on the
seven checkpoints, its sole nonzero rate is Qwen3-14B under the list
instruction, 0.015625, and it reads 0.0000 on all seven translation cells.
On \code{Qwen3-4B-Instruct-2507} the argmax is instead the opening of a response
that carries the instruction out, \code{'Keywords'} in 17 of 64 items under the
list instruction and \code{'Note'} in 30 of 64 under translation, so an
opening-word match and execution of the instruction measure different
properties. The token-class rule predicts an answer letter for the item's own
instruction, a digit for the empty ending and the numeric instruction and a
prose word for the other four; a check holds when the predicted class is the
unrestricted argmax on more than half the items, and counted in the order of
Table~\ref{tab:crossfamily}, the numbers of failing token-class checks are
1, 5, 3, 2, 5, 5 and 2.

\paragraph{The judges and the hand labels.}
Each judge answers every record with its id and one of the four label names, so
a letter that answers the quoted item cannot be read as a label; for a
substituted instruction, the rubric ranks execution above either answer label
and leaves only answer versus forecast unranked. Grok, Claude Opus 5.5 and
GPT-6-astra label the same 1862 records. Each matches the assigned label on all
twelve synthetic cases, six executing and six not. Each labels non-executing all
126 cross-paired rows, which are judged against an instruction they were not
given, and labels executing all 13 real rows that a separate first-token
heuristic selects, 11 of them from \code{Qwen3-4B-Instruct-2507}. The heuristic accepts translation
openers \code{Le, La, Les, L, Voici, Il, Elle, Dans, Traduction, Un, Une, Ce,
Cette, Lorsque, Quand, Selon} or list headings \code{Keywords, keywords, Mots,
Mot}, after stripping leading token-boundary markers, and rejects digit-opening
translation and list rows. Both instruments use the same first 32 items in the
same order, prompt and ending variant: the continuation begins with the token
probe's argmax on 1566 of 1568 rows, with both exceptions on SmolLM2. Execution
rates count any continuation consisting only of a number as a forecast,
regardless of its judge label, which changes no translation, listing or
summarisation cell; a bare number can satisfy both
\code{Reply with a single integer.} and the forecast question, so the numeric
instruction has no attributable execution rate. Three people labelled 112
continuations: 100 drawn from the experimental rows, covering the seven
checkpoints and the seven endings unevenly (32 list and 25 translate rows, 8 or
9 for each other ending; 7 to 34 rows per checkpoint), and the twelve synthetic
cases, which they were given first and not told apart from the rest. They saw
the note, the quoted item, the ending and the continuation, and not the judge's
label. A second, independent three-person labelling covers the same 100
continuations. Every agreement, like the rates, counts a bare number as a
forecast (Table~\ref{tab:judge-agreement}).

\paragraph{Execution rates, item answers and forecasts.}
Each rate uses 32 items and 64-token greedy continuations. Grok's
translation-execution rates are 0.6562, 0.7500, 0.9688, 0.8438, 0.2188 and
0.4375 on the six checkpoints other than falcon, in Table~\ref{tab:crossfamily}
order, and falcon reads 0.0000; Opus and GPT also read translation execution on
those six and 0.0000 on falcon. The listing instruction reads 0.9062 on
\code{Phi-3-mini-4k-instruct} and at most 0.3750 elsewhere. Summarisation reads
at most 0.1875, on \code{Qwen3-8B}, and counting at most 0.5938, on
\code{Qwen3-14B}. With the item's instruction intact, Grok labels 31, 32, 31,
32, 3, 0 and 21 of 32 continuations as item answers, in
Table~\ref{tab:crossfamily} order, Mistral and falcon at the space-appended ending; with it deleted, it labels 32, 32, 32, 18,
32, 31 and 31 as forecasts. Opus and GPT assign the same labels as Grok on every
Qwen3 row in these two conditions and label all 32 Phi-3 base continuations as
item answers.

\begin{table}[h]
\centering\small
\caption{Agreement on execution labels, on the binary executes-or-not reading
except in the last row, which compares the four-way label. Judge pairs are
compared on the 1568 experimental continuations. Each labelling covers 100
continuations; comparisons with a judge or between labellings use its majority
of three, and the pairwise row and Fleiss' $\kappa$ use the individual labels.}
\label{tab:judge-agreement}
\begin{tabular}{@{}lrll@{}}
\toprule
Comparison & Rows & Agreement & $\kappa$ \\
\midrule
Grok and Opus & 1568 & 1549 & 0.958 (Cohen) \\
Grok and GPT & 1568 & 1532 & 0.922 (Cohen) \\
Opus and GPT & 1568 & 1533 & 0.924 (Cohen) \\
\midrule
First majority and Grok, Opus, GPT & 100 & 99, 98, 99 & \\
Second majority and Grok, Opus, GPT & 100 & 100, 99, 100 & \\
First and second majorities & 100 & 99 & \\
First labelling's three labellers, pairwise & 100 & 99, 99, 100 & 0.983 (Fleiss) \\
Second labelling's three labellers & 100 & & 0.983 (Fleiss) \\
\midrule
First majority and Grok, four-way & 100 & all but 6 & \\
\bottomrule
\end{tabular}
\end{table}

\paragraph{Agreement.}
Grok matches the first
majority on all 99 rows the three label unanimously; the one row it misses is a
majority of two on \code{translate}. It agrees with the human majority on all
45 digit-opening rows in the sample that the first-token heuristic rejects,
including the 19 labelled executing. Both validation sets give identical
executes-or-not labels at 64 and at 128 generated tokens (256 on the 36
cross-paired Mistral and falcon rows), and at 128 tokens translation execution
rises on three of the five checkpoints using the original ending and is
unchanged on two. Judged again in reshuffled batches, the 448 unchanged
continuations of the two checkpoints using the space-appended ending keep their
binary execution labels on all 192 translation, listing and summarisation rows
and on 426 of 448 overall.

\section{What occupies a published self-evaluation readout}
\label{app:lmpolygraph}

\paragraph{A maintained implementation.}
A P(True)-style self-evaluation has this configuration when the quoted item
carries its own answer instruction and the scorer reads a different declared
set. lm-polygraph \citep{fadeeva2023lm}, a maintained library and benchmark
for uncertainty estimation, implements P(True) among its estimators (their Table~1) and, at revision \code{32cdf4a7}, enables it in 23 of
the 30 evaluation configurations it ships. On MMLU in its
\code{simple\_instruct} configuration, the quoted input opens with the
instruction to reply with the selected option's letter only, lists the lettered
options and ends with \code{Answer:}. The estimator appends the model's own
answer after \code{Possible answer:}, asks whether it is \code{(A) True} or
\code{(B) False}, ends with \code{The possible answer is:}, and reads the
probability of the token \code{True} at the start of the assistant turn. We ran
it with the library's own loaders, prompt and answer step on 64 items drawn by
its subsampling (seed 1), in bfloat16 as the library loads, and reproduced its
reported P(True) to a log-space difference of 0 on the Qwen3 checkpoints and
$9.5\times10^{-7}$ on Llama-3-8B-Instruct. As the library renders the hybrid
Qwen3-8B and Qwen3-14B, leaving thinking on, \code{<think>} is the argmax on all
64. The library's answer step samples at each checkpoint's
own generation settings and requires two new tokens, so printed answers include
strings such as \code{B<|im\_start|>}.

\paragraph{Correctness ranking on MMLU.}
Using the library setup above on 500 MMLU items, we label each generated
answer by its first standalone A--D letter against gold; an answer without
such a letter is incorrect. Table~\ref{tab:lmpolygraph-mmlu} compares four
readouts for the same generated answer. Library is $\log P(\code{True})$,
the negative of the PTrue estimator's output. Renormalised is
$P(\code{True})/[P(\code{True})+P(\code{False})]$ at the same scored position.
Moved prefills the assistant turn with the answer stem
\code{The possible answer is (} and reads
$P(\code{A})/[P(\code{A})+P(\code{B})]$, where \code{A} denotes True.
Deleted removes the quoted letter instruction before applying the library
readout. The Qwen3 average weights the three checkpoints without thinking
equally. The answer stem raises the Qwen3-averaged AUROC to 0.632 $[0.595, 0.669]$;
comparing answers within each answer letter, with letter-specific AUROCs
weighted by the number of correct--incorrect pairs, gives a gain of
0.133 $[0.079, 0.187]$. Deleting the letter instruction
raises the library readout's Qwen3-averaged AUROC by 0.085 $[0.042, 0.127]$.
In the metric lm-polygraph uses to compare estimators, the
prediction--rejection ratio (PRR; 0 for random rejection, 1 for the oracle)
\citep{fadeeva2023lm}, the library readout has Qwen3-averaged PRR
0.016 $[-0.092, 0.124]$, and $-0.187$ $[-0.353, -0.013]$ on Qwen3-4B, below
random rejection; the answer stem raises the Qwen3-averaged PRR to
0.407 $[0.320, 0.486]$, a paired gain of 0.390 $[0.253, 0.524]$.
With thinking on, as the library renders Qwen3-8B and Qwen3-14B, the answer
step returns only \code{<think>} and a newline on all 500 items of each, so
the estimator scores a thinking token and no answer is correct.

\paragraph{Mass and split at the library's scored position.}
Write $M=P(\code{True})+P(\code{False})$ and $R=P(\code{True})/M$, so
$\log P(\code{True})=\log M+\log R$. On Qwen3-4B, correct answers tend to come
with a smaller $M$, AUROC 0.353 $[0.302, 0.405]$, and, on the 387 items where an
A--D letter is the most probable token at that position, with a more
probable letter, AUROC 0.638 $[0.573, 0.700]$. $R$ alone ranks correctness at
0.674, but Library carries the $\log M$ term and ranks correct answers below
incorrect ones. The direction of $M$ differs between checkpoints: on
Llama-3-8B-Instruct its AUROC is 0.617 $[0.565, 0.668]$, and Library ranks above
chance.

\paragraph{The floor and the ranking.}
At the library's scored position, the mass of the declared set \code{True},
\code{False} is below the floor on every item on the three Qwen3 checkpoints
without thinking; of the four readouts, only Moved clears the floor at the
median on all three.
At that position an A--D letter is the unrestricted argmax on 387, 496 and
500 of 500 MMLU items on Qwen3-4B-Instruct-2507, Qwen3-8B and Qwen3-14B
without thinking, respectively. Of these, 179, 215 and 129 are C or D,
which the outer \code{(A)}/\code{(B)} labels do not offer. With the quoted
letter instruction deleted, a letter is the argmax on 0, 0 and 3 items.
Moved reads the library's own \code{(A)}/\code{(B)} labels after the stem:
its declared set holds the scored position, with median mass above 0.999 and
398, 479 and 494 of 500 items at or above the floor on the three Qwen3 runs.

Renormalised and Moved both rank correctness on MMLU and TriviaQA
(Tables~\ref{tab:lmpolygraph-mmlu} and~\ref{tab:lmpolygraph-trivia}).
Section~\ref{sec:failure} reports the paired differences between these readouts.
In the forecast task, renormalisation at the original position stays
indistinguishable from chance, 0.4929 $[0.4390, 0.5458]$, while the stem reaches
0.6152 $[0.5602, 0.6686]$ (Table~\ref{tab:consequence});
Section~\ref{sec:repair} gives the paired comparison of the two positions.

Counting as correct the 11 Qwen3-4B answers with no standalone letter but
a gold first letter fused to lowercase text, such as \code{Bropical},
gives a Qwen3-averaged gain of 0.120 $[0.067, 0.173]$.

\paragraph{A wrong printed answer.}
On the 500 MMLU items of Table~\ref{tab:lmpolygraph-mmlu}, we replace
lm-polygraph's printed possible answer with a non-gold letter drawn per item
with a fixed seed. On Qwen3-4B-Instruct-2507, Qwen3-8B and Qwen3-14B, with
thinking disabled on the latter two, a bare A--D letter is the unrestricted
argmax at the library's scored position on 259, 500 and 500 items,
respectively. Among these letter argmaxes, 186, 289 and 233 match the gold
letter, 39, 105 and 117 match the printed letter, and 34, 106 and 150 match
neither. Across all 500 items per checkpoint, the gold letter's median
probability is 0.0626, 0.951 and 0.246, against $4.10\times10^{-7}$,
$6.90\times10^{-6}$ and $1.01\times10^{-5}$ for the printed letter.
On 102, 132 and 80 items, respectively, the unrestricted argmax is the gold
\code{C} or \code{D}: it correctly answers the quoted item, and is neither
the printed answer nor one of the outer \code{(A)}/\code{(B)} labels.

\begin{table}[h]
\centering\footnotesize
\setlength{\tabcolsep}{2pt}
\renewcommand{\arraystretch}{1.15}
\caption{lm-polygraph confidence on 500 shared MMLU items: AUROC for the
checkpoint's own answer being correct under the letter label. Intervals
are 95\% item-bootstrap intervals ($B=10{,}000$), with the same resampled
items across readouts and checkpoints. The readouts are defined above;
median mass is the declared set's at its scored position, shown as
Library / Moved. The Qwen3 average weights checkpoint accuracies and AUROCs
equally. Differences use unrounded AUROCs.}
\label{tab:lmpolygraph-mmlu}
\begin{tabular}{@{}lrrrrrrr@{}}
\toprule
Checkpoint & Accuracy & \shortstack{Median mass\\Library / Moved}
& Library & Renorm. & Moved & Deleted & \shortstack{Moved\\$-$ Library} \\
\midrule
\shortstack[l]{Qwen3-4B-\\Instruct-2507} & 0.724
& \shortstack[r]{$1.8\times10^{-11}$\\1.00}
& \shortstack[r]{0.423\\$[0.369, 0.477]$} & 0.674 & 0.610 & 0.560
& \shortstack[r]{$+0.187$\\$[+0.113, +0.260]$} \\
\shortstack[l]{Qwen3-8B,\\no thinking} & 0.764
& \shortstack[r]{$3.4\times10^{-13}$\\1.00}
& \shortstack[r]{0.474\\$[0.413, 0.536]$} & 0.592 & 0.617 & 0.580
& \shortstack[r]{$+0.143$\\$[+0.057, +0.224]$} \\
\shortstack[l]{Qwen3-14B,\\no thinking} & 0.798
& \shortstack[r]{$1.2\times10^{-10}$\\1.00}
& \shortstack[r]{0.611\\$[0.550, 0.670]$} & 0.715 & 0.669 & 0.623
& \shortstack[r]{$+0.059$\\$[-0.017, +0.134]$} \\
Qwen3 average & 0.762 & n/a
& \shortstack[r]{0.503\\$[0.464, 0.540]$} & 0.660 & 0.632 & 0.587
& \shortstack[r]{$+0.129$\\$[+0.078, +0.181]$} \\
\shortstack[l]{Llama-3-8B-\\Instruct} & 0.614
& \shortstack[r]{$3.3\times10^{-4}$\\0.999}
& \shortstack[r]{0.659\\$[0.609, 0.708]$} & 0.618 & 0.597 & 0.607
& \shortstack[r]{$-0.061$\\$[-0.127, +0.005]$} \\
\bottomrule
\end{tabular}
\end{table}

\paragraph{A second task: TriviaQA.}
We use the library's loader, subsample, answer step and P(True) prompt on 500
TriviaQA items. An answer is correct when its first line, stripped of a leading
\code{Answer:}, matches a gold alias after both are stripped of non-ASCII
symbols and nonspacing marks and passed through the library's normalisation.
The library's label uses the full answer without these extra removals:
257 of Qwen3-8B's answers start with \code{Answer:}, and 20 of Qwen3-4B's
answers end in a non-ASCII symbol. At the scored position, one of the library's
option-label tokens \code{(A} and \code{(B} is the unrestricted argmax on
500, 500 and 463 items on the three Qwen3 checkpoints without thinking and
396 on Llama-3-8B-Instruct. On those items the model writes the library's option label
where the library reads \code{True}. Table~\ref{tab:lmpolygraph-trivia} gives
the masses and AUROCs. In PRR, Library scores $-0.013$ $[-0.071, 0.039]$
on the Qwen3 average and $-0.506$ $[-0.585, -0.433]$ on Qwen3-4B, and Moved
scores 0.756 $[0.700, 0.805]$ on the average. With the library's own correctness
label, the Qwen3-averaged AUROCs are 0.418 for Library and 0.866 for Moved.
With thinking on, the answer step returns only \code{<think>} and a newline
on all 500 items of each of Qwen3-8B and Qwen3-14B.

\begin{table}[h]
\centering\footnotesize
\setlength{\tabcolsep}{2pt}
\renewcommand{\arraystretch}{1.15}
\caption{lm-polygraph confidence on 500 shared TriviaQA items: AUROC for the
checkpoint's own answer being correct under the alias-match label defined above.
Intervals are 95\% item-bootstrap intervals ($B=10{,}000$), with the same
resampled items across readouts and checkpoints. The readouts are defined in
the MMLU comparison; here Deleted removes \code{Answer the following question
as briefly as possible.} from the quoted input. Median mass is the declared
set's at its scored position, shown as Library / Moved. The Qwen3 average weights
checkpoint accuracies and AUROCs equally. Differences use unrounded AUROCs.}
\label{tab:lmpolygraph-trivia}
\begin{tabular}{@{}lrrrrrrr@{}}
\toprule
Checkpoint & Accuracy & \shortstack{Median mass\\Library / Moved}
& Library & Renorm. & Moved & Deleted & \shortstack{Moved\\$-$ Library} \\
\midrule
\shortstack[l]{Qwen3-4B-\\Instruct-2507} & 0.344
& \shortstack[r]{$6.4\times10^{-15}$\\1.00}
& \shortstack[r]{0.185\\$[0.147, 0.226]$} & 0.859 & 0.884 & 0.742
& \shortstack[r]{$+0.699$\\$[+0.636, +0.760]$} \\
\shortstack[l]{Qwen3-8B,\\no thinking} & 0.450
& \shortstack[r]{$7.4\times10^{-12}$\\1.00}
& \shortstack[r]{0.395\\$[0.346, 0.446]$} & 0.868 & 0.865 & 0.689
& \shortstack[r]{$+0.470$\\$[+0.403, +0.535]$} \\
\shortstack[l]{Qwen3-14B,\\no thinking} & 0.518
& \shortstack[r]{$5.1\times10^{-8}$\\1.00}
& \shortstack[r]{0.839\\$[0.803, 0.872]$} & 0.852 & 0.853 & 0.763
& \shortstack[r]{$+0.015$\\$[-0.005, +0.035]$} \\
Qwen3 average & 0.437 & n/a
& \shortstack[r]{0.473\\$[0.448, 0.497]$} & 0.860 & 0.868 & 0.731
& \shortstack[r]{$+0.395$\\$[+0.360, +0.428]$} \\
\shortstack[l]{Llama-3-8B-\\Instruct} & 0.604
& \shortstack[r]{$2.0\times10^{-6}$\\1.00}
& \shortstack[r]{0.664\\$[0.616, 0.712]$} & 0.832 & 0.829 & 0.669
& \shortstack[r]{$+0.164$\\$[+0.116, +0.214]$} \\
\bottomrule
\end{tabular}
\end{table}

\paragraph{Leaving out one checkpoint.}
Dropping any one of the three Qwen3 checkpoints from the average keeps Moved's
gain over the library readout above zero: from $+0.1007$ $[+0.0434, +0.1578]$
to $+0.1648$ $[+0.1024, +0.2266]$ on MMLU and from $+0.2424$
$[+0.2065, +0.2771]$ to $+0.5846$ $[+0.5351, +0.6332]$ on TriviaQA.

\paragraph{The labels at the library's position.}
Reading the prompt's own labels where the library reads \code{True}, without a
stem, renormalises the probabilities of \code{(A} and \code{(B}, each one token
on these four checkpoints. On MMLU, where an answer letter is usually the argmax
at that position, the labels' median mass on the three Qwen3 checkpoints is
$2.5\times10^{-7}$, $2.5\times10^{-10}$ and $2.4\times10^{-12}$ (4B, 8B and
14B), and this readout's AUROC is 0.626,
0.561 and 0.502, averaging 0.563 $[0.520, 0.606]$. Moved minus this readout is
$-0.016$ $[-0.048, +0.016]$, $+0.056$ $[+0.024, +0.088]$ and $+0.167$
$[+0.111, +0.223]$, and $+0.069$ $[+0.042, +0.095]$ on the average. On
TriviaQA, where a label is usually the argmax and the labels' median mass is at
least 0.981 on each Qwen3 checkpoint, the readout averages 0.857 $[0.833, 0.880]$ and Moved minus it is
$+0.011$ $[+0.001, +0.021]$. On Llama-3-8B-Instruct, Moved minus it is $+0.037$
$[+0.002, +0.071]$ on MMLU and $-0.004$ $[-0.023, +0.015]$ on TriviaQA. The same
runs reproduce the stored library readout, Renormalised and Moved on every item.

\paragraph{Four further checkpoints.}
The same test, on the same 500 items per task, runs on \code{Phi-3-mini-4k-instruct},
\code{Mistral-7B-Instruct-v0.3}, \code{falcon-7b-instruct} and
\code{SmolLM2-1.7B-Instruct}. Mistral and SmolLM2 load through the library's
loader; Phi-3 and falcon load through the native \code{transformers} classes,
because their bundled modelling code does not run under the pinned
\code{transformers} 5.17.0. The library-readout check holds exactly on every item.
Across all eight checkpoints, \code{True} and \code{False} hold less than half
the mass at the library's scored position on every TriviaQA item and on every
MMLU item except one of Llama-3-8B-Instruct's; their largest median mass, SmolLM2's, is
0.023 on MMLU and 0.153 on TriviaQA. After the answer stem, the library's
\code{(A)} and \code{(B)} labels hold at least half the mass on every TriviaQA
item on all eight.

\paragraph{A 72B checkpoint.}
The same test runs on \code{Qwen/Qwen2.5-72B-Instruct} at \code{495f393}, through
the library's loader in bfloat16, and the library-readout check holds exactly on
every item. \code{True} and \code{False} hold a median mass of
$9.85\times10^{-7}$ on MMLU and $4.27\times10^{-6}$ on TriviaQA, with item
maxima of $1.5\times10^{-3}$ and $6.7\times10^{-3}$. After the answer stem, the
smallest mass the \code{(A)} and \code{(B)} labels hold on any item of either
task is 0.847.

\paragraph{Single-answer estimator comparison on TriviaQA.}
Our comparison includes every entry in lm-polygraph's
\code{default\_estimators.yaml} at revision \code{32cdf4a7} that uses one
generated answer without additional answer samples, auxiliary models or external
data: fourteen library estimators including PTrue (Library). The other entries are
\code{MaximumSequenceProbability}, \code{Perplexity}, \code{MeanTokenEntropy},
\code{MeanPointwiseMutualInformation},
\code{MeanConditionalPointwiseMutualInformation}, \code{SelfCertainty},
\code{RenyiNeg}, \code{FisherRao}, \code{AttentionScore}, \code{RAUQ} at its two
configured settings, \code{CSL} and \code{BoostedProbSequence}.
The additional estimators use their configured settings and the library's own
calculators on the recorded answer tokens, with confidence the negative of
uncertainty. On the first five items per Qwen3 checkpoint, the replay matches direct
library generation in tokens and log-likelihoods; the attention replay also
matches generation attention and both RAUQ values.
On the Qwen3 average with the alias-match labels of Table~\ref{tab:lmpolygraph-trivia},
Moved has the highest AUROC point estimate among the seventeen readouts:
the fourteen library estimators plus Renormalised, Moved and Deleted.
Library ranks twelfth of the fourteen.
The highest library AUROC is 0.815 (\code{MaximumSequenceProbability});
Moved's paired gain over it is $+0.053$ $[+0.024, +0.082]$. Renormalised is
second, at 0.860, with a paired gain over it of $+0.045$ $[+0.016, +0.074]$.
Moved also has higher PRR than every library estimator at both default maximum
rejection fractions. Its paired gain over the best library estimator is
$+0.083$ $[+0.019, +0.148]$ at 1.0 (\code{MaximumSequenceProbability}) and
$+0.100$ $[+0.028, +0.174]$ at 0.5 (the entropy variant of \code{RAUQ}).
These are paired 95\% item-bootstrap intervals using the same resampled items
across readouts and checkpoints.

\paragraph{Where the configuration occurs.}
Self-evaluation and judging quote by design: P(True) quotes the question it asks about
\citep{kadavath2022language}, and a log-probability judge quotes the instruction
and the outputs it grades. The public pipelines below ship this
configuration in their scorers when the quoted material carries its own
answer instruction and the scorer reads a different declared set.
lm-polygraph enables P(True) in 23 of its 30 configurations, and on MMLU the
quoted input carries the instruction to answer with a letter. AlpacaEval, at
revision \code{cd543a14}, reads log-probabilities over its declared set in 12 of
its 44 annotator configurations, AlpacaEval~2.0's default
\citep{dubois2024length} among them, and at least 14 of the 805 instructions it
quotes carry a closed answer format of their own. Of the 131 OpenCompass dataset
configuration files that define a \code{GRADER\_TEMPLATE} and use an LLM
evaluator at revision \code{c4966665}, 27 quote an item with its lettered
choices and parse the judge's output using its first uppercase \code{A} or
\code{B} character, so a bare letter answering the quoted item is read as a
verdict: \code{A} means correct and \code{B} means incorrect.\footnote{OpenCompass,
\code{opencompass/datasets/generic.py}, lines 63--69 and 31--37, at the
revision above.}
We ran the default MMLU judge configuration\footnote{OpenCompass,
\code{opencompass/configs/datasets/mmlu/mmlu\_llmjudge\_gen\_f4336b.py}, at the
revision above, with greedy decoding.} with Llama-3-8B-Instruct as the judge
on 500 MMLU items, each with one correct and one incorrect candidate answer:
1{,}000 calls in the shipped order. The grading prompt prints the gold letter.
The rotation contrast uses the 500 item--candidate pairs in which reordering
moves that letter between the wrong verdict's letter and C or D. The judge
returns the wrong verdict on 44 calls in the former position and 1 in the
latter. In the shipped order it makes 23 errors. On shipped-order calls whose
gold letter is the wrong verdict's letter, errors exceed the rate on the other
calls with the same candidate correctness by about 19, 82\% of the 23.
Every shipped-order response parses as a valid verdict.
In lm-evaluation-harness, the six PISA tasks judged by a language model quote
the item's lettered options with the answer they grade and map every stripped
response other than \code{1} to \code{0}, so a letter is scored as
incorrect,\footnote{lm-evaluation-harness,
\code{lm\_eval/tasks/pisa/utils.py}, lines 291--329.} while none of the 8,412
declared-set tasks \citep{lmevalharness} quotes an item with its own answer
instruction before asking a different question. The checks of
Section~\ref{sec:repair} belong wherever a scorer quotes what it scores.

\section{Published specifications, cell by cell}
\label{app:harness}

The four specifications are measured on 64 items per cell.

\begin{table}[h]
\centering
\small
\caption{Declared-set first-token mass as declared / with a leading space,
$10^{\operatorname{median}(\log_{10}\ctxmass)}$ over 64 items with linear
interpolation on the log scale. G-Eval, Zhuang and Kadavath are completion
prompts ending at \code{- Coherence:}, \code{Output:} and
\code{The proposed answer is:}; AlpacaEval uses each checkpoint's chat template.
Kadavath's question is the item's stem without its lettered choices or
instruction, and the column scores the shared first token of \code{' (A)'} and
\code{' (B)'}; their existing space makes the prefixed variant undefined.
G-Eval's prefixed options share a first token on all seven checkpoints;
Zhuang's and AlpacaEval's have distinct first tokens on all 64 items on every
checkpoint. AlpacaEval's context ends at whitespace except on Mistral and falcon.}

\label{tab:harness}
\begin{tabular}{@{}lrrrr@{}}
\toprule
Checkpoint & G-Eval & Zhuang & AlpacaEval & Kadavath \\
\midrule
Qwen3-4B-Instruct-2507 & \num{1.26e-05} / 0.99449 & \num{1.73e-06} / 0.99684 & 1.000 / \num{2.88e-14} & 0.48600 / --- \\
Qwen3-8B & 0.00783 / 0.96250 & \num{6.04e-04} / 0.91395 & 1.000 / \num{1.07e-10} & 0.82539 / --- \\
Qwen3-14B & 0.00336 / 0.99567 & \num{3.76e-05} / 0.76455 & 1.000 / \num{2.22e-15} & 0.67639 / --- \\
Phi-3-mini-4k-instruct & 0.0174 / 0.65558 & 0.11343 / 0.56764 & 0.94021 / 0.94021 & 0.82375 / --- \\
Mistral-7B-Instruct-v0.3 & \num{2.83e-04} / 0.90844 & \num{4.83e-04} / 0.99172 & \num{9.22e-06} / 0.99426 & 0.39998 / --- \\
falcon-7b-instruct & 0.00606 / 0.90917 & \num{1.31e-04} / 0.96491 & \num{7.01e-04} / 0.0249 & 0.54744 / --- \\
SmolLM2-1.7B-Instruct & 0.0108 / 0.76328 & \num{8.59e-05} / 0.0262 & 0.42963 / \num{6.36e-05} & \num{1.39e-04} / --- \\
\bottomrule
\end{tabular}
\end{table}

\paragraph{The four specifications, and how far the boundary carries.}
AlpacaEval~2.0 is the only one of the four rendered
through a chat template, so the only one whose scored context moves with the
checkpoint and the only one that can discriminate: four of the five checkpoints whose contexts end at whitespace hold 0.94021 or
more, \code{SmolLM2-1.7B-Instruct} ends at whitespace and still reads 0.42963,
and the two whose contexts do not end at whitespace hold $7.01\times10^{-4}$ or
less, and on those two the declared set is absent from
the five candidates the published parser reads in every one of the 64 items.

Zhuang et al.'s named checkpoint, Flan-UL2, and the Flan-T5-XL and XXL
comparators have median decoder sequence masses of 0.99904, 0.99883 and 0.99864,
respectively, with an option argmax on every item and identical bare and
space-prefixed targets; these are sequence probabilities at a decoder reopened
from \code{<pad>}, a different quantity from Table~\ref{tab:harness}'s
first-token masses. Within Table~\ref{tab:harness}, the counterexample is one
cell of the 21 that are scorable: AlpacaEval~2.0 on \code{SmolLM2-1.7B-Instruct} ends at whitespace with
distinct first tokens and an applicable metric, so the rule places it on the clean
side, and its median is 0.42963, beneath the floor. On these cells, the boundary
condition is necessary but not sufficient.
The other seven of the 28, Kadavath's column, are not scorable, because its
two options share a first token on all seven checkpoints.

\paragraph{G-Eval's chat rendering.}
The G-Eval repository's system-message rendering falls below the floor
on all four checkpoints with final whitespace, distinct option first tokens
and an applicable first-token metric: Qwen3-4B, Qwen3-8B, Qwen3-14B and SmolLM2.

\paragraph{The repair's mechanism, in the same harness.}
\code{lm-evaluation-harness} provides assistant prefilling through
\code{gen\_prefix} (\code{lm\_eval/config/task.py:39}), documented as
``Prefix to start the assistant's response'' at \code{lm\_eval/api/task.py:961}.
On its chat path, \code{build\_qa\_turn} creates an assistant message containing
the prefix (\code{lm\_eval/api/task.py:1104-1106}), and
\code{add\_generation\_prompt=not gen\_prefix} suppresses a second assistant
opening (\code{lm\_eval/api/task.py:970}). The Hugging Face backend continues
that final message (\code{lm\_eval/models/huggingface.py:1723-1728}).
Our implementation instead appends the stem after rendering the generation
prompt, as Appendix~\ref{app:prompt} shows. Both place scoring after assistant
content; byte-identical rendering across implementations was not tested.
Nine task files set \code{gen\_prefix}, among them
\code{discrim\_eval\_explicit} and \code{discrim\_eval\_implicit}, whose prefixes
end in a quote character and so carry the boundary their empty
\code{target\_delimiter} would otherwise have dropped. We take the mechanism and
supply a string; what this paper adds is where the failure sits inside its scope
and how far into it the repair reaches (Table~\ref{tab:crossfamily}).

\section{Shared first tokens on four published prompts}
\label{app:geval-dimensions}

Appendix~\ref{app:harness} reads one G-Eval prompt. This reads all four of that
instrument's dimensions on the same seven checkpoints, 100 SummEval \citep{fabbri2020summeval} documents and 1600 rows per cell, 28
cells in all. The four are coherence, consistency, fluency and relevance. \emph{Sign convention in this appendix}: $\Delta\rho =
\rho(\text{bare first token}) - \rho(\text{space-prefixed sequence})$, so a positive value means
G-Eval's own readout agrees with the human ratings better than the repaired one.
$\rho$ is the summary-level Spearman correlation, taken within each source
document across its 16 system summaries and then averaged over the documents
where it is defined; each interval is a paired bootstrap of 10\,000 resamples
with the document as the unit, both scoring methods recomputed on the same resample.

\paragraph{Constant scores and defined correlations.}
Space-prefixed first-token scores are constant on all 28 cells, each with
1600 rows and zero documents with a defined rank correlation for that score.
G-Eval's original options have distinct first tokens on all 28 cells;
their correlations range from $-0.0598$ to $+0.6306$.

\paragraph{Argmax at G-Eval's own scored position.}
At G-Eval's own scored position, across all four dimensions and all seven
checkpoints, the argmax is one of the declared options on \textbf{0 of 44800}
rows.

\paragraph{The 28 differences.}
$\Delta\rho$ moves in both directions, from $-0.2194$ to $+0.0831$; 19 of the 28
are negative and 9 positive, and 14 bootstrap intervals exclude zero.

\paragraph{What the denominators mean.}
The document counts are 100, 96 and 98, decided per document. The 42 missing documents are 28 on consistency and 14 on fluency
across the seven checkpoints. All are missing because the human rating vector is
constant there, so no rank correlation exists; none is a missing measurement and
none is due to a constant model-score vector.

\section{Whether the original position's ordering carries the forecast}
\label{app:consequence}

The target is one deterministic note-only reader outcome per passage and
checkpoint on the 240 RACE-H passages of Table~\ref{tab:consequence}.
Each of Qwen3-4B, Qwen3-8B, Qwen3-14B and Meta-Llama-3-8B-Instruct writes
its own note, reads it and forecasts its own reader outcome. The 960 rows
contain 160, 163, 171 and 140 correct reader answers, respectively.

\paragraph{Writer and reader.}
The writer receives the passage without its question:
\begin{Verbatim}[fontsize=\small,breaklines=true,breakanywhere=true,frame=single]
Source text:
{context}

A future reader will answer a question about this text using ONLY your note. You do not get to see the question.
Budget: at most 32 words.
Write only the note and then stop.
\end{Verbatim}
Generation is greedy with a cap of 256 new tokens. Notes end at an
end-of-sequence token, contain 1 to 32 words and contain no thinking token.
An over-budget or unfinished attempt is retried up to four times. Retry $r$,
counted from 1, requests at most $\lfloor32/(r+1)\rfloor$ words in one compact
phrase. The note-only reader receives one user turn:
\begin{Verbatim}[fontsize=\small,breaklines=true,breakanywhere=true,frame=single]
Note:
{note}

{question}
A: {option A}
B: {option B}
C: {option C}
D: {option D}
Answer with exactly one of: A, B, C, D.
\end{Verbatim}
The checkpoint's chat template opens the assistant turn, with thinking
disabled and no system turn or answer stem. The reader chooses the option
among \code{" A"}, \code{" B"}, \code{" C"} and \code{" D"} with the highest
full-sequence log-probability (float16 forward pass, float32 log-softmax).
The target $y$ is 1 exactly when this option equals gold. The constructed-note
check in Appendix~\ref{app:reader} compares the reader's spaced and unspaced strings
on 240 passages with constructed notes, half stating the answer,
on Qwen3-4B: both pick the same option on 96.25\% of rows.

\paragraph{Forecast and scored outcome.}
The same checkpoint receives the note and quoted item in the forecast prompt
printed in Appendix~\ref{app:prompt}. The phrase \code{Out of 100 such readers}
is the question's wording; the readout is scored as a probability of the
single binary reader outcome, using AUC and Brier.
Readouts (a) and (b) hold the model, dtype, GPU, note, item, declared set and
renormalisation fixed and move the scored position by adding the
answer stem; the five options have the same first-token IDs at both positions
on all four checkpoints. At the production position every passage is
below the mass floor on each Qwen3 checkpoint: the median mass is
$1.54\times10^{-16}$, $1.49\times10^{-15}$ and $1.71\times10^{-11}$, and an
answer letter is the argmax on 240, 233 and 239 of the 240 passages on
Qwen3-4B, 8B and 14B; with the answer stem each median exceeds 0.9999.
Each forecast probability is $p_S$
from Eq.~\ref{eq:forecast-probability}, scoring the full option sequence with its
tokens encoded in isolation and appended to the rendered context; on all
four checkpoints, those tokens equal the in-context tokens.
AUC is ROC AUC against the note-only reader's binary
correctness, counting ties as one half; bootstrap intervals are the 2.5th and
97.5th percentiles of passage-bootstrap replicates (seed 0).
Brier is the mean of $(p_S-y)^2$, where $y$ is reader correctness.
An average is the mean of its checkpoints' values: Table~\ref{tab:consequence} averages the
three Qwen3 checkpoints, and Table~\ref{tab:consequence-readouts} gives that average and
the average of all four.
An AUC difference is the readout minus its reference; a Brier difference is
the reference minus the readout, so positive values favour the readout.

Table~\ref{tab:consequence} in Section~\ref{sec:repair} reports these readings.

\begin{table}[h]
\centering\small
\caption{Constant-forecast Brier baselines on the same evaluation rows as
Table~\ref{tab:consequence}. The fitted constants use each checkpoint's
empirical reader-correctness rate; the Qwen3 average is the mean of its three checkpoints.
These are descriptive baselines fitted on the evaluation rows.}
\label{tab:constant-baselines}
\begin{tabular}{@{}lrrrrr@{}}
\toprule
Forecast & Qwen3-4B & Qwen3-8B & Qwen3-14B & Qwen3 average & Llama-3-8B-Instruct \\
\midrule
Constant 0.5 & 0.2500 & 0.2500 & 0.2500 & 0.2500 & 0.2500 \\
Fitted constant & 0.2222 & 0.2179 & 0.2048 & 0.2150 & 0.2431 \\
\bottomrule
\end{tabular}
\end{table}

The stemmed position's Qwen3-averaged and Llama Brier scores each exceed both
descriptive constant baselines.

The original position's Qwen3-averaged $\mathrm{AUC}$ interval contains 0.5,
so its item-level ranking is not distinguishable from chance.

In Table~\ref{tab:consequence}'s Qwen3 rows, the two positions select the same
grid value on
4.6\%, 9.2\% and 12.9\% of rows, against 96.67\%, 96.25\% and 98.33\% for the reader's spaced and unspaced options on the same rows (Appendix~\ref{app:reader}); the original position puts
\code{10} on 223, 236 and 187 of its 240 rows, while the repaired one spreads
across the grid.

\paragraph{Five readouts on the four checkpoints.}
Table~\ref{tab:consequence-readouts} compares (a) the production prompt without
a stem; (b) that prompt followed by \code{Out of 100 such readers, about };
(c) that prompt with the quoted item's closing instruction deleted, without a
stem; (d) Appendix~\ref{app:general}'s T1 with the original ending,
without a stem; and (e) T1 followed by \code{Estimate: }.
The note and reader target are fixed across readouts, and all intervals use the
same 20,000 passage-cluster bootstrap replicates across readouts and checkpoints.
The Qwen3 checkpoints are read here a second time, on other hardware than
Table~\ref{tab:consequence}'s Qwen3 rows. The second reading differs from
Table~\ref{tab:consequence} in Qwen3-averaged AUC by $-0.0002$ for (a)
and $+0.0004$ for (b) at the printed precision, by at most 0.0011 on a single
checkpoint, and in Brier by at most 0.0002. Llama-3-8B-Instruct was read at snapshot \code{8afb486c},
whose weights, tokenizer and configuration files are the same blobs as those of
Appendix~\ref{app:general}'s \code{e1945c40}: of the 14 files both cached
snapshots hold, only the repository's model card differs.
Deleting the instruction returns the median option mass above 0.9998 on each
Qwen3 checkpoint but not on Llama-3-8B-Instruct, whose median is 0.0023. Against the
stem, the averaged Brier score is worse for (c), (d) and (e), each interval
below zero.

\paragraph{The number the model writes.}
Readout (f) generates greedily from (a)'s prompt, at most 48 new tokens, and
reads the first maximal digit run that is one of the five options, skipping any
other number; the forecast is that value
divided by 100, and a generation with no grid number is scored 0.5. A grid number
is found on 239, 239, 240 and 116 of the 240 passages on Qwen3-4B, Qwen3-8B,
Qwen3-14B and Llama-3-8B-Instruct, whose other 124 generations answer the
question in prose. The forecasts concentrate on one value: 50 on 167 and 221 of
Qwen3-4B's and Qwen3-8B's passages, 70 on 232 of Qwen3-14B's. Their AUCs are
0.5775, 0.5146, 0.4974 and 0.5529, averaged over the four 0.5356 $[0.5083, 0.5628]$;
against (a) the averaged difference is $+0.0353$ $[-0.0155, +0.0857]$ and against (b)
$-0.0772$ $[-0.1169, -0.0355]$. On the Qwen3 rows with a grid number, against (b),
it is $-0.0847$ $[-0.1304, -0.0378]$. The averaged Brier score of (f), 0.2448, is
lower than (b)'s by 0.0089 $[-0.0040, +0.0222]$ and higher than the fitted
constant's 0.2220: a forecast held near one value scores near a constant on
Brier without ranking the passages.

\begin{table}[!htbp]
\centering\footnotesize
\setlength{\tabcolsep}{2pt}
\renewcommand{\arraystretch}{1.1}
\caption{Five forecast readouts on the same 240 passages per checkpoint.
The upper panel gives AUC with its 95\% interval and Brier for the Qwen3
and four-checkpoint averages, each with equal checkpoint weights.
Paired differences in the lower panel use the four-checkpoint average and
$\Delta\mathrm{AUC}=\mathrm{AUC}(x)-\mathrm{AUC}(r)$ and
$\Delta\mathrm{Brier}=\mathrm{Brier}(r)-\mathrm{Brier}(x)$, so positive values favour $x$.
All intervals use a paired passage-cluster bootstrap shared by every readout and checkpoint.
Readouts (a)--(e) are defined in the text above the table.}
\label{tab:consequence-readouts}

\begin{tabular}{@{}lrrrr@{}}
\toprule
& \multicolumn{2}{c}{Qwen3 average} & \multicolumn{2}{c}{Four-checkpoint average} \\
Readout & AUC [95\% interval] & Brier & AUC [95\% interval] & Brier \\
\midrule
(a) & 0.4927 [0.4388, 0.5457] & 0.4748 & 0.5003 [0.4544, 0.5452] & 0.4221 \\
(b) & 0.6156 [0.5607, 0.6689] & 0.2543 & 0.6128 [0.5634, 0.6605] & 0.2538 \\
(c) & 0.5651 [0.5083, 0.6218] & 0.2713 & 0.5640 [0.5148, 0.6124] & 0.2644 \\
(d) & 0.5926 [0.5381, 0.6460] & 0.2930 & 0.5879 [0.5404, 0.6342] & 0.2800 \\
(e) & 0.6015 [0.5445, 0.6570] & 0.2933 & 0.6065 [0.5564, 0.6550] & 0.2823 \\
\bottomrule
\end{tabular}
\par\smallskip
\begin{tabular}{@{}llrr@{}}
\toprule
Reference $r$ & Readout $x$ & $\Delta\mathrm{AUC}$ [95\% interval] & $\Delta\mathrm{Brier}$ [95\% interval] \\
\midrule
(a) & (b) & $+0.1125\, [0.0477, 0.1766]$ & $+0.1683\, [0.1412, 0.1951]$ \\
(a) & (c) & $+0.0637\, [0.0056, 0.1218]$ & $+0.1577\, [0.1347, 0.1800]$ \\
(a) & (d) & $+0.0876\, [0.0243, 0.1501]$ & $+0.1421\, [0.1204, 0.1628]$ \\
(a) & (e) & $+0.1062\, [0.0422, 0.1709]$ & $+0.1398\, [0.1181, 0.1610]$ \\
\midrule
(b) & (a) & $-0.1125\, [-0.1766, -0.0477]$ & $-0.1683\, [-0.1951, -0.1412]$ \\
(b) & (c) & $-0.0489\, [-0.0791, -0.0191]$ & $-0.0106\, [-0.0202, -0.0007]$ \\
(b) & (d) & $-0.0250\, [-0.0518, 0.0019]$ & $-0.0263\, [-0.0418, -0.0104]$ \\
(b) & (e) & $-0.0063\, [-0.0291, 0.0160]$ & $-0.0286\, [-0.0422, -0.0145]$ \\
\bottomrule
\end{tabular}
\end{table}

\begingroup
\raggedbottom
\paragraph{Short answer stems.}
On the three Qwen3 checkpoints, the forecast is also read after \code{Number: }
and after \code{Forecast: }, in the same run as
Table~\ref{tab:consequence-readouts}'s readouts (a) and (b), whose values it
reproduces. Each stem is three tokens at the end of the rendered context and
leaves the five options' first tokens unchanged. After either stem the option
mass has a median of at least 0.997 on each checkpoint, with at most 6 of 240
passages below the floor. Readout (b) reaches AUC 0.6286, 0.6446 and 0.5738 on
Qwen3-4B, 8B and 14B; \code{Number: } reaches 0.6052, 0.6620 and 0.5763, and
\code{Forecast: } 0.6082, 0.6461 and 0.5674. Averaged over the three,
\code{Number: } is $+0.1218$ $[+0.0500, +0.1935]$ above readout (a) and
$-0.0011$ $[-0.0213, +0.0191]$ from readout (b); \code{Forecast: } is
$+0.1145$ $[+0.0414, +0.1872]$ above (a) and $-0.0084$ $[-0.0328, +0.0162]$
from (b).

\section{The reading under eight prompt conditions}
\label{app:general}

\paragraph{The probe and what is held fixed.}
The probe repeats Table~\ref{tab:crossfamily}'s measurement on its seven
checkpoints at the snapshots in Appendix~\ref{app:snapshots}, and on
\code{meta-llama/Meta-Llama-3-8B-Instruct} at \code{e1945c40}, loaded from a cache on the
compute host.
Each condition uses 64 items per checkpoint: the same first 64 RACE-H items
in frozen order, except for D1's item replacement. The five forecast options
remain \code{10}, \code{30}, \code{50}, \code{70}, \code{90}.
Every reading uses a float16 forward pass, float32 log-softmax and the
checkpoint's own chat template through its generation prompt, with
\code{enable\_thinking=False} where the template reads that argument.
All eight conditions were read on one GPU model.

C0 reads the production prompt again, with Table~\ref{tab:crossfamily}'s
fixed note, \code{Landlord only gives when told to take, not give.}
The user-turn forecast text and the quoted item's closing letter instruction
are those printed in Appendix~\ref{app:prompt}; C0 changes that appendix's
example note to this fixed note. N1 changes the note, S1 the turn containing
the forecast instruction, and D1 the dataset and item format. A1 and A2 insert
sentences into C0, and T1 and T2 replace its whole user-turn template.
The fixed note is used in every condition except N1.

The mass without an answer stem is read at the original ending. The answer stem
is \code{Out of 100 such readers, about\ } for C0, N1, S1, D1, A1 and A2.
T1 and T2 are read with their own answer stems, printed below, and also with
the original answer stem. On Mistral and falcon,
whose templates omit final whitespace, the separator control appends
one space to the rendered prompt. Each prompt also receives a greedy
continuation of at most 16 new tokens from the original ending.

Each condition uses the collapse definition in Section~\ref{sec:protocol}
at the original ending; letter pull means that at least half
the items have one of their own answer letters as the unrestricted argmax;
repair means a median at or above the floor with the condition's primary
answer stem. The letter count strips whitespace and a tokeniser's leading
word-boundary marker. Letter pull excludes Llama-3-8B-Instruct's \code{A}
when its continuation begins \code{A clever}, \code{A classic} or \code{A nice}:
these are prose openings. Table~\ref{tab:mechanisms} separates the mechanisms.

\paragraph{N1: per-item notes.}
N1's notes are the ones \code{Qwen3-4B-Instruct-2507} wrote for these 64 items
in a separate reading with a twelve-word budget;
the same 64 notes replace C0's fixed note for every checkpoint. All other
prompt text is C0's.

\paragraph{S1: system and user turns.}
The system text is exactly:
\begin{Verbatim}[fontsize=\small,breaklines=true,breakanywhere=true,
                 breaksymbolleft={\tiny\ensuremath{\hookrightarrow}},frame=single,
                 samepage=true]
You will be given a note and a quoted question. A future reader will be shown ONLY the note, not the source text, and will then be asked the quoted question. Out of 100 such readers, how many answer correctly? Reply with exactly one of: 10, 30, 50, 70, 90.
\end{Verbatim}

The user text is exactly the following template, with \code{\{note\}}
replaced by the fixed note and \code{\{q\}} by the unchanged RACE-H item
block, including its closing answer instruction:
\begin{Verbatim}[fontsize=\small,breaklines=true,breakanywhere=true,
                 breaksymbolleft={\tiny\ensuremath{\hookrightarrow}},frame=single,
                 samepage=true]
Note:
{note}

Quoted question:
{q}
\end{Verbatim}

Mistral's chat template merges the system text into the user turn on all
64 items. The other seven templates retain it in a separate block on all
64 items each; all eight therefore have S1 readings.

\paragraph{D1: MMLU-Pro items.}
D1 uses the first 64 rows of the MMLU-Pro test split in stored order
\citep{wang2024mmlupro}. There are 62 items with ten
options, one with nine and one with seven. The item block replacing
\code{\{q\}} has exactly this form:
\begin{Verbatim}[fontsize=\small,breaklines=true,breakanywhere=true,
                 breaksymbolleft={\tiny\ensuremath{\hookrightarrow}},frame=single,
                 samepage=true]
{question}
{rows}
Answer with exactly one of: {letters}.
\end{Verbatim}

Here \code{\{question\}} is the row's question, \code{\{rows\}} is its
options joined by single newlines, each as \code{<letter>: <option text>},
and \code{\{letters\}} is that row's letters joined by comma-space.
The letters start at A and end at the item's last option. The forecast
template and fixed note remain C0's.

\paragraph{A1 and A2: added sentences.}
A1 inserts \code{The question\textquotesingle{}s instruction to answer with a letter is for
the readers.} and one following space immediately before
\code{Out of 100 such readers, how many answer correctly?} in C0.
A2 adds \code{Your task is to forecast their number of correct answers.}
and one following space immediately after the first insertion. Both sentences
come from T1 in that order; the answer stem and all other text remain C0's.

\paragraph{T1 and T2: whole forecast templates.}
A language model, \code{gpt-6-astra}, wrote both templates from a description of the forecasting
task and a made-up example item, with no information about the failure or
the checkpoints; that task text is printed below in full. It was asked for two complete user-turn templates with
\code{\{note\}} and \code{\{q\}} placeholders and an answer stem for each.
The only checks were mechanical: the JSON parsed, each placeholder occurred
once, there were no other placeholders, and neither stem contained a digit.
Both templates were used as written. In each, \code{\{note\}} is the fixed
note and \code{\{q\}} is the unchanged RACE-H item block. Each assigns the
letter instruction to the readers and ends with an output contract.
As in Appendix~\ref{app:prompt}, arrows in these boxes mark typeset line
breaks rather than newlines in the prompt.

\begin{samepage}
The task given to the writer, in full:
\begin{Verbatim}[fontsize=\scriptsize,breaklines=true,breakanywhere=true,
                 breaksymbolleft={\tiny\ensuremath{\hookrightarrow}},frame=single,
                 samepage=true]
# Write two prompt templates

We need prompts for the following forecasting task, to be sent as the user message to
several open instruction-tuned chat models through each model's own chat template.

The situation: someone read a text passage and wrote a short note about it (a sentence or
two). Later, a group of readers will be shown ONLY that note, never the passage, and will
then answer one multiple-choice question about the passage. The question is shown to them
exactly as it appears in the block below, including the answer instruction at its end.

The model we prompt sees the note and that same question block. It must forecast how many
out of 100 such readers answer the question correctly, and reply with exactly one of the
five values 10, 30, 50, 70, 90.

A question block looks like this (a made-up example; real ones come from a
reading-comprehension dataset and always have this shape):

    What does the writer mainly want to tell us?
    A: Travel is expensive for students.
    B: Small towns are worth visiting.
    C: Trains are faster than buses.
    D: Maps are often out of date.
    Answer with exactly one of: A, B, C, D.

Write TWO templates for this task, written the way you would write them for real use.
Make them differ from each other in structure and wording, not only in small phrasing.
Requirements:
- Use the placeholders `{note}` for the note text and `{q}` for the question block. `{q}`
  must appear exactly once and be inserted unchanged; do not rewrite or reformat it.
- Use no other placeholders. Plain text only.
- The template is the whole user message. Do not add a system message.
- The reply the model gives must be exactly one of 10, 30, 50, 70, 90.
- For each template, also give an answer lead-in: the first words of an ideal assistant
  reply, ending right before the number (for example `Estimate: `). It must end with a
  space or a colon-space and must not contain a number.

Write the result to `templates.json` in this directory, as
`[{"id": "T1", "template": "...", "answer_lead_in": "...", "notes": "..."}, {"id": "T2", ...}]`,
and check that the file parses as JSON and that each template contains `{note}` and `{q}`
exactly once and no other `{...}`.
\end{Verbatim}
\end{samepage}

\begin{samepage}
T1, in full:
\begin{Verbatim}[fontsize=\small,breaklines=true,breakanywhere=true,
                 breaksymbolleft={\tiny\ensuremath{\hookrightarrow}},frame=single,
                 samepage=true]
Someone read a passage and wrote the short note below. Imagine 100 readers who will see only this note and the question below, including its final answer instruction. They will never see the passage. Each reader will answer the question, and correctness will be judged against the original passage.

Note:
{note}

Question:
{q}

Estimate how many of these 100 readers will answer correctly. Consider how much useful information the note provides, whether it distinguishes the choices, and where readers may misunderstand or have to guess. The question's instruction to answer with a letter is for the readers. Your task is to forecast their number of correct answers.

Choose the closest of these five counts to your estimate: 10, 30, 50, 70, 90. Reply with exactly one of these values and nothing else.
\end{Verbatim}

T1's answer stem is \code{Estimate:\ }.
\end{samepage}

\begin{samepage}
T2, in full:
\begin{Verbatim}[fontsize=\small,breaklines=true,breakanywhere=true,
                 breaksymbolleft={\tiny\ensuremath{\hookrightarrow}},frame=single,
                 samepage=true]
OUTPUT CONTRACT
Return one bare value from this set: 10, 30, 50, 70, 90.

FORECAST TARGET
Number of correct answers among 100 readers taking a reading-comprehension test. Each reader receives only a short note written by someone who read the passage, plus the complete test item below. Neither you nor the readers receive the passage. The passage's answer key determines correctness.

TEST ITEM SHOWN TO READERS
BEGIN TEST ITEM
{q}
END TEST ITEM

NOTE AVAILABLE TO READERS
BEGIN NOTE
{note}
END NOTE

ASSESSMENT PROCEDURE
1. Identify what information the item asks readers to recover from the passage.
2. Assess which choices the note supports or rules out. Account for missing or ambiguous information, plausible mistakes, and guessing.
3. Estimate how many readers will select the correct answer, then select the nearest permitted count.

Keep the assessment to yourself. The letter-answer instruction inside the test item governs the readers' answers. Your response must follow the output contract: exactly one of 10, 30, 50, 70, 90, with no label, explanation, punctuation, or percent sign.
\end{Verbatim}

T2's answer stem is \code{Predicted number correct:\ }.
\end{samepage}

\paragraph{The readings.}
Tables~\ref{tab:general} and~\ref{tab:mechanisms} are computed from the per-item
in-context masses and argmax tokens; the letter-pull count also reads the
opening of each Llama continuation. Medians are displayed to five decimal places
or four significant figures in scientific notation; counts use the
unrounded values. T1 and T2 remove letter pull on every checkpoint; the three
below-floor cells in each are Mistral's and falcon's omitted separators, plus
Phi-3's shared first token under T1 and SmolLM2 under T2. With the stem,
Llama-3-8B-Instruct's C0 median is 0.99352.

\begingroup
\footnotesize
\sisetup{group-digits=false}
\setlength{\tabcolsep}{2pt}
\begin{longtable}{@{}lrrrrrrrr@{}}
\caption{Eight prompt conditions on 64 items per checkpoint. Each cell gives
the raw in-context median at the original ending, followed by the letter-argmax
count and below-floor count, separated by a semicolon; both counts are out of 64.
$\dagger$ marks shared first tokens. Llama's counted letters are the article
\code{A}. Under S1, Mistral's template merges the system text into the user
turn. Checkpoint names
are abbreviated from Table~\ref{tab:crossfamily}, with Llama-3-8B-Instruct added.
With each condition's own answer stem, every median is at or above the floor
except falcon's under D1, 0.45421, and T2, 0.33207.}
\label{tab:general}\\
\toprule
Checkpoint & C0 & N1 & S1 & D1 & A1 & A2 & T1 & T2 \\
\midrule
\endfirsthead
\multicolumn{9}{l}{Table \thetable\ continued}\\
\toprule
Checkpoint & C0 & N1 & S1 & D1 & A1 & A2 & T1 & T2 \\
\midrule
\endhead
\bottomrule
\endlastfoot
\shortstack[l]{Qwen3-4B\\\strut} & \shortstack[r]{\num{2.707e-15}\\63;64} & \shortstack[r]{\num{2.669e-16}\\64;64} & \shortstack[r]{\num{0.00404}\\48;53} & \shortstack[r]{\num{2.761e-10}\\61;64} & \shortstack[r]{\num{5.824e-07}\\56;61} & \shortstack[r]{\num{0.99991}\\5;9} & \shortstack[r]{\num{1.00000}\\0;0} & \shortstack[r]{\num{1.00000}\\0;0} \\[3pt]
\shortstack[l]{Qwen3-8B\\\strut} & \shortstack[r]{\num{8.144e-14}\\61;64} & \shortstack[r]{\num{1.183e-14}\\61;64} & \shortstack[r]{\num{0.93951}\\25;25} & \shortstack[r]{\num{3.107e-11}\\14;64} & \shortstack[r]{\num{0.00900}\\42;50} & \shortstack[r]{\num{0.99895}\\1;2} & \shortstack[r]{\num{1.00000}\\0;0} & \shortstack[r]{\num{0.99963}\\0;0} \\[3pt]
\shortstack[l]{Qwen3-14B\\\strut} & \shortstack[r]{\num{1.371e-08}\\62;64} & \shortstack[r]{\num{4.235e-11}\\63;64} & \shortstack[r]{\num{1.404e-04}\\60;61} & \shortstack[r]{\num{2.718e-11}\\39;64} & \shortstack[r]{\num{5.807e-12}\\62;64} & \shortstack[r]{\num{0.00218}\\54;60} & \shortstack[r]{\num{1.00000}\\0;0} & \shortstack[r]{\num{1.00000}\\0;0} \\[3pt]
\shortstack[l]{Phi-3-mini\\\strut} & \shortstack[r]{\num{5.801e-04}$^{\dagger}$\\63;64} & \shortstack[r]{\num{9.051e-04}$^{\dagger}$\\61;64} & \shortstack[r]{\num{0.00137}$^{\dagger}$\\64;64} & \shortstack[r]{\num{0.00271}$^{\dagger}$\\18;64} & \shortstack[r]{\num{0.00169}$^{\dagger}$\\63;64} & \shortstack[r]{\num{0.03618}$^{\dagger}$\\61;64} & \shortstack[r]{\num{0.39024}$^{\dagger}$\\20;46} & \shortstack[r]{\num{0.53902}$^{\dagger}$\\2;26} \\[3pt]
\shortstack[l]{Mistral v0.3\\\strut} & \shortstack[r]{\num{3.488e-06}\\30;64} & \shortstack[r]{\num{1.907e-06}\\31;64} & \shortstack[r]{\num{5.023e-07}\\57;64} & \shortstack[r]{\num{5.412e-07}\\17;64} & \shortstack[r]{\num{6.074e-06}\\3;64} & \shortstack[r]{\num{4.694e-06}\\0;64} & \shortstack[r]{\num{2.005e-06}\\0;64} & \shortstack[r]{\num{1.180e-07}\\0;64} \\[3pt]
\shortstack[l]{falcon\\\strut} & \shortstack[r]{\num{1.051e-04}\\0;64} & \shortstack[r]{\num{1.274e-04}\\0;64} & \shortstack[r]{\num{3.158e-04}\\0;64} & \shortstack[r]{\num{6.909e-05}\\0;64} & \shortstack[r]{\num{1.108e-04}\\0;64} & \shortstack[r]{\num{9.853e-05}\\0;64} & \shortstack[r]{\num{1.385e-04}\\0;64} & \shortstack[r]{\num{6.051e-05}\\0;64} \\[3pt]
\shortstack[l]{SmolLM2\\\strut} & \shortstack[r]{\num{0.39017}\\46;53} & \shortstack[r]{\num{0.41918}\\41;52} & \shortstack[r]{\num{0.01977}\\64;64} & \shortstack[r]{\num{0.08264}\\57;62} & \shortstack[r]{\num{0.48962}\\23;34} & \shortstack[r]{\num{0.65508}\\1;1} & \shortstack[r]{\num{0.82232}\\0;0} & \shortstack[r]{\num{0.31081}\\7;64} \\[3pt]
\shortstack[l]{Llama-3-8B\\\strut} & \shortstack[r]{\num{2.565e-04}\\49;64} & \shortstack[r]{\num{1.113e-04}\\8;64} & \shortstack[r]{\num{0.58728}\\0;21} & \shortstack[r]{\num{2.626e-04}\\47;64} & \shortstack[r]{\num{1.141e-05}\\46;64} & \shortstack[r]{\num{1.128e-05}\\49;64} & \shortstack[r]{\num{0.95180}\\0;0} & \shortstack[r]{\num{0.84624}\\0;0} \\[3pt]
\end{longtable}
\endgroup

\begin{table}[h]
\centering\small
\setlength{\tabcolsep}{3pt}
\caption{What holds the scored position at the original ending in each condition of
Table~\ref{tab:general}, counted over its eight checkpoints. \emph{Answer letters}:
one of the item's own letters is the unrestricted argmax on at least 32 of 64 items,
not counting Llama-3-8B-Instruct's article \code{A}. \emph{Prose opening}: a word
token that opens prose, including \code{Answer} and that article, is the argmax on
at least 32. \emph{Omitted separator}: the rendered ending lacks the whitespace the
options need. \emph{Shared first token}: the options' in-context first tokens
coincide. \emph{Options}: median mass at or above the floor with distinct first
tokens. Rows below the first can overlap. The last row counts repairs.}
\label{tab:mechanisms}
\begin{tabular}{@{}lrrrrrrrr@{}}
\toprule
 & C0 & N1 & S1 & D1 & A1 & A2 & T1 & T2 \\
\midrule
Median below the floor & 8 & 8 & 6 & 8 & 8 & 5 & 3 & 3 \\
\midrule
Answer letters & 5 & 5 & 5 & 3 & 4 & 2 & 0 & 0 \\
\quad separator present, first tokens distinct & 4 & 4 & 3 & 3 & 3 & 1 & 0 & 0 \\
Omitted separator & 2 & 2 & 2 & 2 & 2 & 2 & 2 & 2 \\
Shared first token & 1 & 1 & 1 & 1 & 1 & 1 & 1 & 1 \\
Prose opening & 1 & 1 & 0 & 4 & 1 & 1 & 0 & 0 \\
Below the floor, none of the above & 0 & 0 & 0 & 0 & 1 & 0 & 0 & 1 \\
Options hold the position & 0 & 0 & 2 & 0 & 0 & 3 & 5 & 4 \\
\midrule
Repaired by the condition's own stem & 8 & 8 & 8 & 7 & 8 & 8 & 8 & 7 \\
\bottomrule
\end{tabular}
\end{table}

\paragraph{C0 compared with Table~\ref{tab:crossfamily}.}
C0 retains all three classifications on the original seven checkpoints.
Qwen3-8B, Qwen3-14B, Mistral and falcon were first read on C0's GPU model
and agree item by item. Qwen3-4B, Phi-3-mini and SmolLM2 were first read
on another GPU model; their medians change by at most 0.19\% and their
minima by at most 3.3\%, and SmolLM2's below-floor count changes from
50/64 in Table~\ref{tab:crossfamily} to 53/64. All letter rates and stemmed
below-floor counts agree.

\paragraph{A larger checkpoint.}
\code{Qwen/Qwen3-32B} at snapshot \code{9216db57} was read under C0 and in the
instruction experiment of Section~\ref{sec:failure}, on another GPU model than the
seven checkpoints of Table~\ref{tab:crossfamily}. At
the production position its median in-context mass is $3.79\times10^{-3}$
(minimum $5.91\times10^{-7}$), all 64 items are below the floor, and no item has
a forecast option at the argmax: the argmax is an answer letter on 35 items
(rate 0.547) and \code{Answer} on 29. With the stem the median is 0.99928
(minimum 0.87448) and no item is below the floor. Deleting the imperative takes
the letter rate to 0.000 and the median to 0.99085, with a digit at the argmax
on all 64. Of its C0 continuations, 48 answer the item first, 13 of them after
\code{Answer:}, and the other 16 open with \code{Answer:} followed by a number
(Table~\ref{tab:c0-continuations}). The token-class prediction holds for the
prose and numeric substitutes and fails for the list, translation and count
substitutes, where a digit takes the position on 38, 49 and 64 of 64 items.

\paragraph{What the unrestricted continuation writes.}
Table~\ref{tab:c0-continuations} classifies the C0 continuations: the
production prompt with the fixed note and the original ending, no answer stem,
greedy decoding of at most 16 new tokens. Fixed rules on the opening, regular expressions and string comparisons with
the item's question and options, assign the first class that matches: an answer
to the item, a forecast, a recital of the option list, the question or an
instruction, other prose, or other. A continuation
that opens as prose and then answers the item, such as
``\code{Based on the note, the correct answer is D:}'', ``\code{D is not true, as
the note}'' or ``\code{The text is mainly about B:}'', counts as prose; a separate
flag, not tabulated, for an answer letter anywhere in the text catches the first
and the third. Text after a special token or a new chat turn is dropped. When a
continuation stops at the token limit after a single line labelled with an
item letter, the rules cannot tell an answer from the first line of an option
list, and they count it as an answer. A forecast that comes after a long answer
can also be cut off, so both flag columns are lower bounds, most visibly on Phi-3-mini (27 of 63).
Three people labelled 100 of the 512 continuations of the eight checkpoints
other than Qwen3-32B by what each does first,
with the checkpoint and the rule class hidden. The sample is stratified by rule
class and checkpoint and contains the three continuations quoted above; on ten
practice records with fixed answers each annotator scored 10 of 10. They agree
with each other on 99 of the 100 (Fleiss' $\kappa=0.991$), and their majority
label matches the rule class on 92, or on 97.9\% when the sample is reweighted
to all 512. Seven of the eight differences are continuations the rules class as
prose and all three annotators label an answer to the item: the three quoted
above and four falcon continuations that state an option's content as the
answer. The eighth is a falcon line labelled with an item letter at the token
limit, which the rules count as an answer and the annotators as the start of a
recited list. None of the eight is on a Qwen3 checkpoint. By the annotators'
reading, the rules undercount answers to the item.

\begin{table}[h]
\centering
\small
\setlength{\tabcolsep}{3pt}
\caption{Classes of the C0 continuations, greedy, at most 16 new tokens, on the
same 64 items for each checkpoint. The first five columns partition the items
by the opening; the last two are flags that overlap them. ``Answer, then
grid'': the opening answers the item and a grid value follows it. ``Grid
anywhere'': a grid value anywhere in the text, outside a recited option list.
A continuation that gives a forecast first can give a value off the grid.}
\label{tab:c0-continuations}
\begin{tabular}{@{}lrrrrrrr@{}}
\toprule
Checkpoint & \shortstack{Answers\\first} & \shortstack{Forecast\\first}
& \shortstack{List or\\question} & Prose & Other
& \shortstack{Answer,\\then grid} & \shortstack{Grid\\anywhere} \\
\midrule
Qwen3-4B-Instruct-2507 & 63 & 1 & 0 & 0 & 0 & 63 & 63 \\
Qwen3-8B & 63 & 0 & 0 & 1 & 0 & 61 & 61 \\
Qwen3-14B & 62 & 0 & 0 & 2 & 0 & 62 & 62 \\
Phi-3-mini-4k-instruct & 63 & 0 & 0 & 1 & 0 & 27 & 27 \\
Mistral-7B-Instruct-v0.3 & 29 & 30 & 0 & 5 & 0 & 2 & 28 \\
falcon-7b-instruct & 3 & 35 & 18 & 8 & 0 & 0 & 35 \\
SmolLM2-1.7B-Instruct & 46 & 18 & 0 & 0 & 0 & 0 & 18 \\
Meta-Llama-3-8B-Instruct & 4 & 1 & 0 & 59 & 0 & 0 & 1 \\
Qwen3-32B & 48 & 16 & 0 & 0 & 0 & 38 & 54 \\
\bottomrule
\end{tabular}
\end{table}

\paragraph{The continuations and the counted tokens.}
Every letter counted for Meta-Llama-3-8B-Instruct is the article in
\code{A clever question!}, \code{A classic} or \code{A nice}. These are prose
openings, rather than answers to the item.

Phi-3-mini's raw mass sums a shared first token (id \code{29871}) in
every condition and item. Those masses are not separability readings,
including its T1 collapse. Its stemmed first tokens are distinct on every
item in all eight conditions. SmolLM2 under T2 holds a raw median of
0.31081, below the floor on all 64 items, with an option at the argmax
on 56/64.

\paragraph{The omitted separator.}
Mistral and falcon collapse under all eight conditions at the original ending.
With the separator supplied, Mistral's median exceeds the floor in
all eight. Falcon's medians with the separator supplied are
\num{0.38482}, \num{0.39438}, \num{0.69728}, \num{0.33924}, \num{0.34499}, \num{0.37506}, \num{0.44811}, \num{0.14991}
in the condition order used above, and its option-argmax counts are
63, 63, 64, 61, 63, 63, 62, 40, all out of 64.
Thus an option is at falcon's argmax on most items in every condition,
while its median remains below the floor in seven; S1 clears it.

\paragraph{The original answer stem on T1 and T2.}
For comparison with their own stems, T1 and T2 are also read with
\code{Out of 100 such readers, about\ }. This gives eight repairs on T1 and
seven on T2, with falcon below the floor on T2, as with the templates' own
stems.
No item has an item-letter or A--D answer-letter argmax with the primary
stem in any of the eight conditions on the eight checkpoints, with the
original answer stem on T1 and T2, or in the additional Qwen3-32B C0 reading.
\par
\endgroup

\end{document}